\documentclass{article}
\pdfinfoomitdate=1 \pdftrailerid{} \pdfsuppressptexinfo=-1
\usepackage{iclr2027_conference,times}
\usepackage{url}\usepackage{booktabs}\usepackage{graphicx}\usepackage{amsmath,amssymb}\usepackage{xcolor}\usepackage{array}
\usepackage{float}
\usepackage[tableposition=top]{caption}
\usepackage{placeins}
\usepackage{pdflscape}
\usepackage{hyperref}\hypersetup{hidelinks}

\usepackage{amsmath,amsfonts,bm}

\def\eqref#1{equation~\ref{#1}}

\def\1{\bm{1}}

\DeclareMathAlphabet{\mathsfit}{\encodingdefault}{\sfdefault}{m}{sl}
\SetMathAlphabet{\mathsfit}{bold}{\encodingdefault}{\sfdefault}{bx}{n}

\newcommand{\vLexSigProxyfrozenMistral}{12}

\newcommand{\vLexSigRflenientQwen}{0}

\newcommand{\vRuSigParentQwen}{8}
\newcommand{\vRuSigParentMistral}{8}
\newcommand{\vRuSigExtraQwen}{3}
\newcommand{\vRuSigExtraMistral}{12}
\newcommand{\vRuSigBundleQwen}{8}
\newcommand{\vRuSigBundleMistral}{12}

\newcommand{\vRuBatchASFQwenEZero}{13}
\newcommand{\vRuBatchAFQwenEZero}{13}

\newcommand{\vRuBatchASFQwenEFour}{11}
\newcommand{\vRuBatchAFQwenEFour}{13}

\newcommand{\vRuBatchASFMistralEZero}{4}
\newcommand{\vRuBatchAFMistralEZero}{10}
\newcommand{\vRuBatchADiscMistralEZero}{0/6}

\newcommand{\vRuBatchAHolmMistralEZero}{0.125}
\newcommand{\vRuBatchASFMistralEFour}{4}
\newcommand{\vRuBatchAFMistralEFour}{10}
\newcommand{\vRuBatchADiscMistralEFour}{1/7}

\newcommand{\vRuBatchAHolmMistralEFour}{0.211}

\newcommand{\vRuBatchBMinHolmStrict}{0.562}

\newcommand{\vRuEchoRealListsN}{36}
\newcommand{\vRuEchoRealListsPassSevenOne}{34}
\newcommand{\vRuEchoRealListsPassOld}{28}

\newcommand{\vRvItemChanged}{0}
\newcommand{\vRvStrictChanged}{0}

\newcommand{\vRuMemFailStrict}{332}
\newcommand{\vRtMemFailStrict}{347}

\hypersetup{pdftitle={When Evidence Changes: Evaluating Memory Repair and Re-reading in Language-Model Agents},pdfauthor={Wenhui Chu}}
\title{\raggedright When Evidence Changes: Evaluating Memory Repair and Re-reading in Language-Model Agents}
\author{Wenhui Chu\\
Department of Cybersecurity\\
College of Emergency Preparedness, Homeland Security and Cybersecurity\\
University at Albany, State University of New York\\
Albany, NY 12222, USA\\
\texttt{wchu3@albany.edu}}
\iclrfinalcopy
\begin{document}
\maketitle
\lhead{Preprint}
\begin{abstract}
When documents supporting an agent's derived facts are revoked or replaced, should it repair memory or re-read current evidence? We introduce an evidence-revision evaluation on medication- and problem-list tasks from public ICU records. Under revocation, replacement and control events, we compare full and source-filtered re-reading with caching, rebuilding and graph-local repair across two 7B models. Memory is supplied in full without retrieval, and costs include ingest, revision and every use. On short records, local repair uses 5--10$\times$ fewer revision tokens than rebuilding, yet every memory pipeline costs at least twice full re-reading in held-out conditions. In a small pre-specified development sweep, adding task-ineligible documents extended records to about 10,000 tokens; at that length, memory's mean cumulative cost fell below full re-reading's after 2--14 uses, partly through truncated extraction, while source-filtered re-reading remained cheapest. In the replacement study, none of the four primary confirmatory tests reached statistical significance. These results show why the cost of agent memory after evidence revision must be assessed against source-filtered re-reading over the full pipeline.

\end{abstract}

\section{Introduction}\label{sec:intro}
A records assistant ingests a pharmacy record, extracts a medication, and writes a summary; later the connector revokes the record as misfiled. The raw record is marked revoked; the extracted claim and the summary are not, and in a new session the assistant builds a handoff list \emph{from memory}. Nothing in this chain is an injection: the source was authentic when written and every write-time check passed. After a well-defined evidence revision, when should an agent maintain and repair derived memory, and when is re-reading the current evidence the better pipeline? We treat it as an evaluation question with three axes reported together: correctness under the revision, retention of still-valid information, and full pipeline cost, including initialisation, repair and every use.

Agent memory systems persist derived memory \citep{chhikara2025mem0,ji2026infini}, and recent work shows that such memory can keep serving outdated or revoked facts \citep{chao2026stale,shen2026revoked}. Existing studies answer adjacent questions: repair systems score repair quality and operator cost against exhaustive repair \citep{zhao2026memorepair}, maintained memory has been compared with full context on accuracy \citep{patel2026supersede}, and cost studies price memory against long context or full transcripts without revision-triggered repair \citep{pollertlam2026beyond,pollertlam2026totalrecall} or, concurrently, meter memory operations and full-transcript replay \citep{mishra2026merit}. To our knowledge, none charges ingest, revision and every use against re-reading the current records, with and without source filtering, after a document is revoked or replaced (\S\ref{sec:related}).

Under full-pipeline accounting, full re-reading was cheaper than every memory pipeline on the short records studied. \emph{Cost:} graph-local repair's revise phase cost 5--10$\times$ fewer tokens than rebuilding's, yet its complete pipeline used 3.5--4.3$\times$ re-reading's tokens on held-out patients, and reuse did not close the gap (\S\ref{sec:cost}). \emph{Length:} when a pre-specified sweep lengthened the same records to about 10{,}000 tokens with chart documents the task does not use, the mean cumulative cost of all three memory pipelines fell below full re-reading's after 2--14 uses, partly through truncated extraction; re-reading only the allowed sources stayed cheapest at every length, and even task-filtered memory used 1.4--1.5$\times$ as many tokens per use (\S\ref{sec:length}). \emph{Replacement:} re-reading adopted a replaced field for 13/13 development patients per model against 8--9/13 for rebuild and local repair, but every miss omitted a new value that was in the memory prompt, and most were lines already omitted before the revision; no 7B comparison of success (adoption with every other medication kept) between re-reading and these pipelines was significant, and none of the four primary tests of a pre-specified confirmatory replacement run (C2) rejected (\S\ref{sec:replace}). \emph{Revocation:} no pipeline listed a revoked drug in 288 uses, as expected once the revoked document and derived nodes are removed, and a pre-specified held-out test at the small size did not detect an excess of non-allowed drugs under memory (\S\ref{sec:revoke}).

This paper contributes (i) an evidence-revision protocol: allowed sources per task, a support class for every medication key, revocation, replacement and control events with expected answers, and endpoints for stale use, retention, adoption and cost (\S\ref{sec:task}); (ii) full-pipeline accounting that charges ingest, revision and every use, with a standard break-even decomposition $\Delta C(k)$ and a source-filtered re-reading baseline, applied to memory pipelines that share extraction, a per-document versioned dependency graph and a read-set commit gate (\S\S\ref{sec:method},~\ref{sec:cost}); (iii) evidence from two open 7B models: development batches, a pre-specified held-out run, the confirmatory run C2 and a pre-specified length sweep, each labelled by evidential role in Table~\ref{tab:batchesfull} (App.~\ref{app:repro}); and (iv) a scorer-sensitivity audit of the problem-list task under lexical rules fixed after outputs were seen (\S\ref{sec:measure}).

Our threat model is narrow: an authentic connector revokes, replaces or adds documents; we model no adversary who writes documents, alters memory or injects instructions. Source permissions are stated in the use prompt and scored, not enforced (only two pipelines of \S\ref{sec:length} filter by source). The commit gate checks only that nothing an action read changed between snapshot and commit, not support, permission or freshness of rendered nodes (no-propagation committed superseded values), and never blocked a commit here. Zero counts are observations, not guarantees; record fidelity is not clinical validity; the ordering is a whole-pipeline result, not a mechanism attribution.

\section{Related Work}\label{sec:related}
\paragraph{Memory maintenance and repair in agents.}
MemoRepair \citep{zhao2026memorepair} treats source deletion, correction and migration as repair events, withdraws the cascade reached through influence edges, and selects repaired successors to republish by a fixed-$\lambda$ value--cost objective (min-cut); its value and cost terms are fixed inputs, and it gates only repaired successors, measuring downstream actions separately. Dependency-guided rollback repair \citep{yu2026rollback} takes a failed trace and a diagnosed fault set, preserves nodes with independent support, and replays only answer-relevant steps; Infini Memory \citep{ji2026infini} consolidates memory into revisable topic documents and EvoMem \citep{xu2026evoarena} records patch-based update histories. \citet{he2026stored} define typed provenance with supersession and revocation and mediate assertions at release, tested on 24 hand-authored cases without a language model. MemSecBench \citep{chen2026memsecbench} follows poisoned memory through a Write--Execute--Forget protocol with repair checkpoints. These works study how memory is maintained or repaired; we ask whether maintaining it is worth its full cost against re-reading.

\paragraph{Provenance, read-set validation and transactional commit.}
MemLineage \citep{ouyang2026memlineage} signs entries, records a derivation DAG, verifies entries at retrieval and gates sensitive actions on lineage; TMA-NM \citep{louck2026tmanm} binds origin at write time with a machine-checked separation theorem. CaMeL \citep{debenedetti2025camel} and FIDES \citep{costa2025fides} enforce capability or information-flow policies at tool calls, evaluated on prompt injection in AgentDojo \citep{debenedetti2024agentdojo}. Their endpoints are attack success and task utility. EAL-Bench \citep{cerruti2026eal} shows that misrecorded permissions and revocations in memory can grant authority the history never permitted, without any attacker. Cordon \citep{chen2026cordon} validates a transaction over tool intents, lineage and staged effects before commit; MemTX \citep{li2026memtx} stages memory writes in snapshot-isolated transactions, blocks irreversible tool calls while uncommitted beliefs are in flight and repairs derived records when a belief is retracted; MemTxn \citep{cui2026memtxn} admits an update only if its value is in the cited evidence and restores the application-visible state after faults; \citeauthor{khan2026concurrency}~\citeyearpar{khan2026concurrency} formalises the stale-generation anomaly and verifies runtimes that prevent it; PlanFence \citep{chen2026planfence} has plans cite the records they used and, before an external action, validates only the records that can affect it, replanning or blocking when validation is incomplete. Our read-set commit gate is closest to these designs: it rejects commits whose read set changed and enforces no source-support, provenance, permission or information-flow policy.

\paragraph{Attacks, stale memory and record-level evaluation.}
Memory poisoning is studied by AgentPoison \citep[including an electronic-health-record agent]{chen2024agentpoison}, MPBench \citep{dash2026mpbench} and GhostWriter \citep{torres2026ghostwriter}; these adversarial writes are outside our threat model. STALE \citep{chao2026stale} reports a gap between retrieving updated evidence and acting on it; \citet{shen2026revoked} find that no memory system of five enforces revocation in every case, over direct and extraction-based writes; the two that returned both the revoked record and its replacement (Graphiti by default, mem0 only with its default read-time filter overridden) ranked the revoked record first; StateAuditor \citep{sun2026stateauditor} checks the provenance and chronology of state transitions before draft repair; TEPA \citep{zhou2026tepa} makes validity a revocable state of evidence memory; and Supersede \citep{patel2026supersede} measures update accuracy lost when bounded self-maintained memory replaces full context.

\paragraph{Cost of memory against long context and replay.}
\citet{pollertlam2026beyond} price a Mem0-based fact memory \citep{chhikara2025mem0}, written once and read at a roughly fixed cost per turn, against long-context turns under prompt caching; memory became cheaper after about ten turns at 100k tokens, the direction of our length sweep at a tenth of that length. Their follow-up \citeyearpar{pollertlam2026totalrecall} meters three memory systems' ingest, retrieval and answer calls against full-transcript resubmission (break-even from turn 0 to never within 400 turns); neither repairs memory nor filters re-reading by source. Concurrent work (MERIT; \citealt{mishra2026merit}) meters every memory operation of tool-using agents, probes updated facts and finds full-transcript replay never the most cost-effective condition; our re-reading reads only current records and was cheaper than every memory pipeline on short ones. \citet{wu2026invalidation} evict stale cached API-recovery suggestions by version stamp; row-level invalidation recovered 29--33\,\% of the token cost of a baseline that re-derives every episode, on four of seven models. \citet{spencer2026groundtruth} finds full-history reading tied or beat memory at a short horizon, while at a long one a layered memory at about half the read cost was not reliably less accurate. AgingBench \citep{zhu2026agingbench} diagnoses four kinds of agent aging, including revision.

\paragraph{What this paper adds.}
Cascade propagation, cost-aware repair, support preservation and commit-time validation are established mechanisms, and we claim no novelty for them (an adaptation of the MemoRepair selector is in Apps.~\ref{app:adapted} and~\ref{app:batchE}). We add the evidence-revision protocol, full-pipeline accounting against re-reading with and without source filtering, and the scorer-sensitivity audit.

\section{Problem and Task Definition}\label{sec:task}
\paragraph{Setting.} An agent serves a patient episode rendered as connector documents $D_1,\dots,D_n$, each with a connector-assigned source (\texttt{pharmacy}, \texttt{ehr} for electronic health record, \texttt{lab}), patient identifier, version and payload digest. A memory pipeline derives nodes (claims, i.e.\ extracted facts; per-document summaries; a global summary) from the documents; a re-reading pipeline derives nothing. A \emph{revision event} changes one document version. Afterwards the agent answers use tasks; each answer is an action that a commit gate validates against the prompt's snapshot.

\paragraph{Tasks and allowed sources.} Two use tasks follow every revision (the reuse and length sweeps of \S\S\ref{sec:cost} and~\ref{sec:length} issue only the medication task): a handoff \emph{medication list}, which may use only \texttt{pharmacy} documents, and an active \emph{problem list}, which may use only the problem-list document. Only the currently active version of an allowed document supports an item; patient attribution is the connector's field, never model text. Every pipeline's use prompt states the allowed sources in the same words (a task contract hashed before any scenario existed).

\paragraph{Support classes.} For every medication key of the allowed documents' \texttt{MEDICATION} lines the builder records its support set and a class: \emph{unique-support-revoked} (must be absent after the revision), \emph{still supported} (must remain), \emph{conflict} (replacement), \emph{irrelevant} (only non-allowed documents mention it; listing it is an unsupported-source error, not a stale error), or \emph{ambiguous} (combination components, excluded from expected sets). The drug named by E4's added record (below) is on no allowed line, so its key is added explicitly as irrelevant.

\begin{figure}[t]\centering\includegraphics[width=\textwidth]{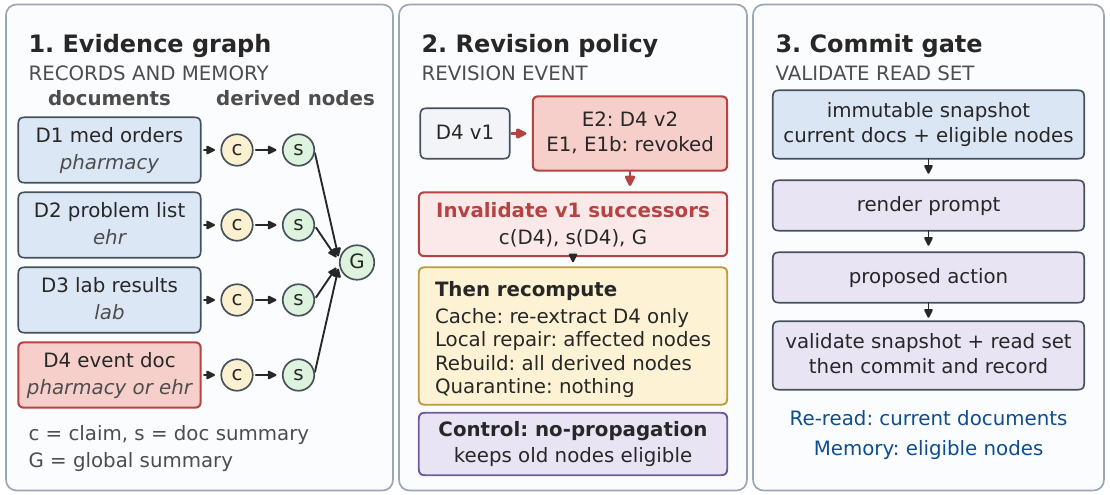}
\caption{The shared harness at the small size. D1--D3 are the patient's medication orders, problem list and laboratory results, with their sources; only \texttt{pharmacy} documents are allowed for the medication task. D4 is the event document: a donor patient's pharmacy record revoked as misfiled after the first use (E1, E1b), a pharmacy order replaced by a version with one changed strength (E2), or an added non-allowed treatment record (E4, \texttt{ehr}); E0 has no revision. Ingest runs one extraction call per document; each claim node (c) records the document version it was shown as its parent, and a per-document summary (s) and a global summary (G) complete the graph. A revision invalidates exactly the transitive successors of the revised version, except under the no-propagation control, which keeps old nodes eligible; the others differ in what they recompute (\S\ref{sec:method}). Every use starts from an immutable snapshot that supplies the prompt and the read set validated at commit.}\label{fig:pipeline}\end{figure}
\paragraph{Revision events and controls.} \textbf{E0}: no revision. \textbf{E1}: a fourth authentic-source document carrying a real medication of a donor patient is revoked as misfiled after the first use. \textbf{E1b}: as E1, but the drug retains independent allowed support and must be retained. \textbf{E4}: no revision; a non-allowed treatment record naming an extra drug is added, and the answer must not change. \textbf{E2}: one pharmacy order is replaced by a second version in which exactly one explicit strength field is changed (a counterfactual $2\times$ change chosen by a pre-declared rule from lines that carry a strength; never a clinical claim); the new value must be adopted and every other valid medication retained.

\paragraph{Endpoints (over attempted trajectories; execution failures stay in the denominator).} Medication use: \emph{strict completion} (listed generic-key set equals the expected set), \emph{stale commit} (a unique-support-revoked key listed), \emph{unsupported-source} and \emph{fabricated} listings, \emph{missing} keys, \emph{retention} of the still-supported key. Replacement: \emph{new value adopted}, \emph{old value retained}, \emph{missing}, \emph{other identities retained}, \emph{success} = adopted $\wedge$ other identities retained (identity level; an unsupported extra identity does not fail it---the stricter \emph{success$_{\text{strict}}$} that excludes it is reported for C2 and the 32B supplement, Apps.~\ref{app:batchCtwo} and~\ref{app:largemodel}). Problem list: the frozen leaf-match proxy, the record-fidelity rules of \S\ref{sec:measure} and their relaxations, side by side. Cost: calls and prompt+completion tokens per phase.

\paragraph{Statistics.} The patient is the unit, except in batch E's unit pairing (App.~\ref{app:batchE}) and one post-hoc pooled sign test (App.~\ref{app:batchCtwo}); conditions, seeds and versions are never pooled. Within-patient comparisons use exact McNemar tests with Holm correction within each declared family (Holm-$m$: a family of $m$ tests); the four C2 primary cells (two models $\times$ two strata) are each tested alone at $\alpha=0.05$, without correction across cells; hypotheses whose direction was fixed in advance (the held-out E4 and C2 primaries, the C2 adoption secondary and their 32B repeats) are one-sided, as is that sign test; all others are two-sided (Apps.~\ref{app:heldout}, \ref{app:batchCtwo} and~\ref{app:largemodel}); proportions carry exact Clopper--Pearson 95\,\% intervals.

\section{Pipelines}\label{sec:method}
All memory pipelines (Fig.~\ref{fig:pipeline}) share the extraction prompt, gateway, node representation and initial ingest; claims extracted from a document that changed before the write are rejected as stale input; validation and the action record form one transaction (App.~\ref{app:harness}). The pipelines are: \textbf{re-read} (no memory; the current documents are placed in the prompt at use time), \textbf{cache} (per-document extraction cache; on revision re-extract the changed document only; no summary layer), \textbf{rebuild} (recompute every node after a revision), \textbf{local repair} (recompute only the invalidated nodes in dependency order: the changed document's claims, its summary and the global summary), \textbf{quarantine} (invalidate the successors and recompute nothing), and \textbf{no-propagation} (called minimal authentication in the frozen plans: historical derived nodes stay eligible for rendering; a negative control, not a candidate design); the last two ran in the replacement batches only. The memory-pipeline use prompt shows all its summaries and facts (no retrieval) with claim identifiers and the parent document's connector source; document versions are tracked by the host read set, not shown to the model. An \textbf{adapted selector} implements a published value--cost repair selector (App.~\ref{app:adapted}); it is an adaptation, not the published method, run once (\S\ref{sec:replace}).

\section{Experimental Setup}\label{sec:setup}
\paragraph{Data.} Episodes come from the intensive care unit (ICU) records of the eICU Collaborative Research Database demo v2.0.1 \citep{pollard2018eicu,johnson2021eicu,pollard2026physionet}, under a person-level split frozen before the experiments reported here (147 development / 54 calibration / 161 sealed people). ``Held-out'' means held out from protocol, prompt and scorer development: no agent output for a sealed patient existed before the held-out run, although the sealed records had been used in an earlier project split (disclosed in the pre-specification). The calibration partition was unused until the confirmatory replacement run of \S\ref{sec:replace}. Each episode is rendered as connector documents: medication orders, a problem list and laboratory results (\emph{small}: 3 documents); the \emph{large} size adds per-day order, laboratory and treatment documents (4--8 documents). Record payloads are short: 65--129 (small) and 210--558 (large) tokens for the eight development screening episodes and 59--155 for the 20 sealed held-out episodes (small only), counted as $1.3\times$ whitespace words; with document headers and tags, development small-size post-revision re-reading bodies are 251--373 Qwen2.5 tokens (App.~\ref{app:length}). The eICU \texttt{medication} table records orders, not confirmed administration \citep{pollard2018eicu}. Nothing is padded.
\paragraph{Models and decoding.} Qwen2.5-7B-Instruct \citep{yang2024qwen25} and Mistral-7B-Instruct-v0.3 \citep{jiang2023mistral,mistral2024instructv03} ran on vLLM \citep{kwon2023efficient} at temperature 0 (seed 0) in every build stage. A labelled parser closes and logs Mistral's truncated JSON (App.~\ref{app:parse}).
\paragraph{Batches (Table~\ref{tab:batchesfull}, App.~\ref{app:repro}).} Development screening and E1b/E4 controls (8 patients, both sizes); one held-out run (20 sealed patients, small size) submitted once and read once; replacement batch C (13 development patients), the shared-ingest batch E (the same 13 plus 4 screening patients) and the confirmatory C2; a same-facts ablation (batch A) and its decoding replay (batch B); a reuse sweep (EXP-R) and a length sweep (EXP-X). Every batch had its code, scenarios and the scoring contract of its pre-declared endpoints hashed before any model output of that batch existed, and every batch called pre-specified also had its analysis plan hashed in advance (``pre-specified'' throughout means this prospective, in-project hash record; no external registry timestamp exists; Table~\ref{tab:batchesfull} gives each batch's role); the problem-list contract and its amendments (\S\ref{sec:measure}) are post-hoc instruments, fixed before re-scoring but after outputs had been seen.

\section{Results}\label{sec:results}
\subsection{Revocation: stale commits, completion, retention and source compliance}\label{sec:revoke}
\emph{Observed stale listings are zero in every cell where the endpoint applies.} In the 288 post-revocation medication uses of the E1 screening and held-out runs (128 development: 8 patients $\times$ 2 sizes $\times$ 4 pipelines $\times$ 2 models; 160 held-out: 20 $\times$ 4 $\times$ 2) no pipeline listed the unique-support-revoked key (E1b has no such key: its drug keeps independent support). Each E1 cell is 0 of 8 or 0 of 20 (exact 95\,\% interval for $0/20$: 0--0.168); cells share patients, so no pooled bound is formed. Later E1 batches (batch E, the length sweep, the 32B runs) had none either. Every pipeline removes the revoked document or its derived nodes before the next use, so this endpoint verifies the implementation rather than separating pipelines; no positive control ran under revocation (no-propagation ran only under replacement, \S\ref{sec:replace}). Strict completion was low at the large size except for Mistral's re-reading (Qwen 2--4 of 8; Mistral re-reading 7 of 8, memory 1--4) and mixed at the small size, mostly through missing orders; memory pipelines also listed drugs from non-allowed records (up to 3 of 8 in E1 cells, 4 in one E0 cell) and, for Mistral, names in no document (up to 3 of 8).

\emph{Retention and source compliance (development, 8 patients per cell).} The still-supported drug was retained in 5--8 of 8 episodes in every E1b cell (App.~\ref{app:heldout}). Qwen's re-reading listed no non-allowed drug in E4 at either size, while its memory pipelines did in 2--4 of 8 at both sizes; Mistral's memory pipelines did in at most 1 of 8 at the small size, and rebuild and local repair in 6 and 5 of 8 at the large size (re-reading 1), where any drug from a non-allowed document counts. In each of the 10 small-size listings (Qwen cache 2, rebuild 3, local repair 4; Mistral local repair 1) the drug had entered memory as a medication fact whose parent is the non-allowed document, as it had in 38 trajectories that did not list it: the path shows how the drug reached the prompt, not why it was listed.

\emph{Pre-specified held-out run (20 sealed patients per model; Table~\ref{tab:heldout}).} Its design, endpoint, test and stop rules were fixed before the agent protocol touched any sealed patient (App.~\ref{app:heldout}). The primary hypothesis---that the memory pipelines list a drug found only in a non-allowed document more often than re-reading---was \emph{not} supported: Qwen 2, 0 and 1 of 20 (cache, rebuild, local repair) against 0 for re-reading (smallest one-sided $p=0.25$); Mistral 1, 4 and 4 against 4, with balanced discordant pairs for rebuild and local repair (3/3 each) and 1/4 for the cache. The run used the small size only, where Mistral's development listings had been rare (at most 1 of 8, against 5--6 at the large size). Later 32B analyses (App.~\ref{app:largemodel}) are exploratory.
\begin{table}[t]\centering
\caption{Held-out run: medication use after the revision or, in E4, the added record (small size; 20 sealed patients per model; counts of 20; retained is shown as a fraction). Stale = unique-support-revoked key listed; retained = still-supported key present; unsupported = non-allowed E4 drug listed; tokens = mean prompt+completion per trajectory. All conditions with intervals: Table~\ref{tab:heldoutfull} (App.~\ref{app:heldout}).}\label{tab:heldout}
\footnotesize\setlength{\tabcolsep}{4pt}\begin{tabular}{@{}llrrrrrrr@{}}\toprule
model & pipeline & \multicolumn{2}{c}{strict completion} & stale (E1) & retained (E1b) & unsupported (E4) & \multicolumn{2}{c}{tokens/traj.}\\
 & & E1 & E4 & & & & E1 & E4\\\midrule
Qwen & re-read & 18 & 18 & 0 & 18/20 & 0 & 1827 & 1911\\
Qwen & cache & 15 & 16 & 0 & 18/20 & 2 & 4129 & 4173\\
Qwen & rebuild & 18 & 18 & 0 & 19/20 & 0 & 9717 & 6958\\
Qwen & local repair & 17 & 17 & 0 & 18/20 & 1 & 7375 & 6971\\
\midrule
Mistral & re-read & 14 & 11 & 0 & 19/20 & 4 & 2018 & 2122\\
Mistral & cache & 13 & 14 & 0 & 18/20 & 1 & 4924 & 4998\\
Mistral & rebuild & 16 & 14 & 0 & 18/20 & 4 & 11607 & 8226\\
Mistral & local repair & 16 & 12 & 0 & 18/20 & 4 & 8676 & 8243\\
\bottomrule\end{tabular}

\end{table}

\subsection{Cost: nodes recomputed versus tokens spent}\label{sec:cost}
Local repair does what it is designed to do: in all 32 E1 screening trajectories it invalidated exactly the three transitive successors of the revoked document and rebuilt one node, where rebuild recomputed 7--14.5 nodes, so its revise phase costs 5--10$\times$ less than rebuild's (Fig.~\ref{fig:cost} and Table~\ref{tab:cost}, App.~\ref{app:costdetail}). But ingest and the memory-based use prompts dominate: in the E1 development cells the complete local-repair pipeline costs 3.96--5.03$\times$ the tokens of re-reading (rebuild 5.3--6.8$\times$, cache 2.2--3.0$\times$) because the extracted facts plus summaries are longer than the raw documents. The held-out run shows the same order: local-repair full-trajectory tokens were 3.5--4.3$\times$ re-reading's over the four held-out conditions (Table~\ref{tab:heldoutphases}, App.~\ref{app:heldout}; use-stage prompts alone 2.1--2.2$\times$), and re-reading was cheapest in every held-out cell. At 32B the ordering is unchanged on the same eight patients: memory cost 2.3--6.5$\times$ re-reading's tokens, with longer later uses in all six cells (App.~\ref{app:largemodel}).
\paragraph{Why reuse does not rescue memory here.}
For $k$ later uses the token difference between a memory pipeline and re-reading is
$\Delta C(k) = \Delta I + \Delta U_0 + \Delta R + \sum_{j=1}^{k}(U_{\mathrm{mem},j}-U_{\mathrm{reread},j})$, where $\Delta I$, $\Delta U_0$ and $\Delta R$ are the ingest, initial-use (pre-revision) and revision differences, together the fixed difference, and $U_{\cdot,j}$ is the token count of later use $j$.
$\Delta I$ and $\Delta R$ are non-negative by construction (re-reading performs no ingest and no revision work); $\Delta U_0$ was non-negative in 11 of 12 cells of the 20-use sweep below. The per-use term decides whether reuse can ever pay. A pre-specified 20-use sweep on the eight development patients (128 trajectories, none failed; App.~\ref{app:reuse}) found it positive in 11 of the 12 model--size--pipeline cells (+175 to +2{,}276 tokens); local repair's later uses cost 1.4--2.5$\times$ re-reading's, and its cumulative ratio fell from 4.9--6.2$\times$ at $k=1$ to 1.8--2.9$\times$ at $k=20$ without reaching 1. The one exception, the Mistral large-size cache ($-12.6$ tokens per use against a fixed difference of 8{,}213; break-even near 652 uses, not observed), had shorter prompts, which in the screening run of that cell came from capped extractions that dropped laboratory content (App.~\ref{app:rho}). No cell's mean cumulative cost fell below re-reading's by $k=20$ (lowest ratio 1.21, 95\,\% interval 0.94--1.47): on these short records graph-local repair saves recomputation, not tokens.
\begin{figure}[t]\centering\includegraphics[width=\textwidth]{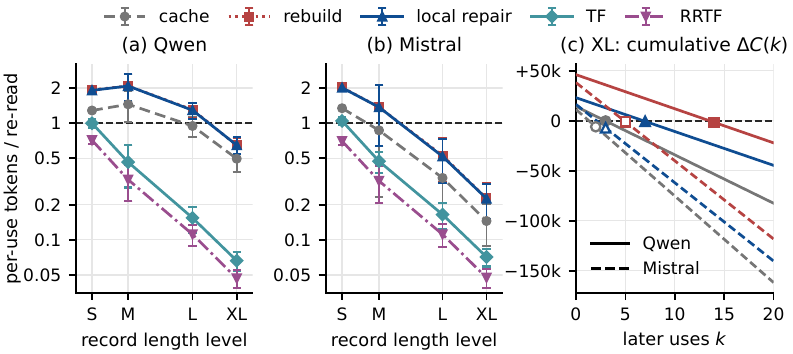}
\caption{Length sweep (E1, 20 later uses; 8, 3, 5 and 6 patients at S, M, L and XL, each level on its own patients). (a, b) Per-use tokens relative to re-reading (mean of per-patient ratios, 95\,\% $t$-intervals), with levels placed at their re-reading tokens per use on a log scale; re-reading used 0.6k--10.1k tokens per use from S to XL. TF, RRTF: task-filtered local repair and re-reading (allowed sources only). (c) Mean cumulative token difference of cache, rebuild and local repair from re-reading at XL; markers at the observed break-even (filled Qwen, open Mistral).}\label{fig:length}\end{figure}
\subsection{Longer records: a pre-specified length sweep}\label{sec:length}
A pre-specified development batch (EXP-X) stress-tested growth in content the task cannot use, with its plan, frozen code, scenarios and analysis program hashed before its first model call (App.~\ref{app:length}). On the eight development screening patients (E1, 20 later uses, both 7B models), each record was lengthened with the same patient's laboratory, vital-sign and nursing-chart documents (whole documents per table and ICU day; nothing padded or synthesised), from sources the medication task does not allow, to four levels of the post-revision re-reading body: the small scenario (S) and targets of 1{,}000 (M), 4{,}000 (L) and 10{,}000 (XL) tokens (Qwen2.5 tokenizer; L is not the large size). A patient entered a level only if its re-reading body fell within 90--100\,\% of the target, giving 8, 3, 5 and 6 patients; each level is analysed on its own patients, and M is descriptive. Two source-filtered pipelines were added: re-reading that shows only allowed-source documents (RRTF), and task-filtered local repair (TF), which still ingests every added document but whose use prompt shows only allowed-source claims and those documents' summaries. All 264 trajectories completed.

\emph{Results against the five pre-specified predictions (Fig.~\ref{fig:length}; App.~\ref{app:length}, Tables~\ref{tab:lengthpred}--\ref{tab:lengthextract}).} At S the per-use difference was again positive for cache, rebuild and local repair in both models (P1). The per-use ratio to re-reading fell from S to XL in all six model--pipeline cells (P2, computed afterwards as the hashed program omitted it; Qwen 1.28--1.93 to 0.50--0.65, Mistral 1.34--2.03 to 0.14--0.23), though not monotonically for Qwen, whose ratios rose from S to M on the same three patients (local repair 1.86 to 2.08) before falling. At XL cache, rebuild and local repair all cost less per use than re-reading, with 95\,\% intervals excluding zero (P3), and cumulative cost crossed within 20 uses: at $k=3$, 7 and 14 for Qwen's cache, local repair and rebuild and at 2, 3 and 5 for Mistral's; at L only Mistral's three crossed (4, 8 and 15). These are crossings of the patient mean; the 95\,\% interval of the 20-use ratio still includes 1 for Qwen's rebuild at XL and Mistral's rebuild at L. Fed the fixed difference and uses 1--3, the identity predicted every observed crossing of a memory pipeline (15 cells, TF included) to within one use (P4), a weak test since per-use cost barely varied over 20 identical uses. Source-filtered re-reading was the cheapest pipeline at every level in both models (P5), at 0.70--0.72$\times$ re-reading's 20-use tokens at S and 0.05$\times$ at XL, because it never pays for the added documents; TF's per-use tokens remained 1.38--1.50$\times$ RRTF's at every level. At L and XL, 225 of the 520 extraction calls on added documents stopped at the 2{,}048-token output cap (none of the 836 on the original documents at any level), and 44 Mistral extractions failed to parse even after repair, so part of memory's per-use saving there is content that never entered memory; complete extractions would have added output and rendered claims, so these break-evens probably favour memory (not measured). P5 was largely implied by the design. Strict completion did not differ significantly from re-reading in any cell (Holm $p=1.0$ throughout), and no trajectory listed the revoked key. Under the pre-specified cached-input pricing (input tokens that repeat the previous call's prefix are billed at a fraction $c$), we report Qwen only: the analysis program counted Mistral's repeated prefixes with the Qwen2.5 tokenizer, so Mistral's cached ratios are not valid and no claim uses them (its no-cache ratios come from server usage and are unaffected; App.~\ref{app:length}). At XL and $k=20$, Qwen's cache and local repair stayed below full re-reading at $c=0.5$ (0.68, 0.92; rebuild 1.13), and none of the three did at $c=0.1$ (1.13--2.37). Caching discounts the record that re-reading resends, not memory's distinct extraction calls. Tokens were weighted equally; pricing output above input would raise memory's fixed cost, since its extractions of added documents averaged about 1{,}500 output tokens each (App.~\ref{app:length}) while re-reading's answers were short (98--118 completion tokens each, held-out run).

\subsection{Replacement (batch C): adoption versus success}\label{sec:replace}
Thirteen development patients, each with one pharmacy order replaced by a counterfactual $2\times$-strength version, were run through five pipelines and both models (130 trajectories; no execution failure; Table~\ref{tab:batchC}, App.~\ref{app:batchC}). Re-reading adopted the new value in 13/13 for both models; rebuild in 8/13 and 9/13; local repair in 9/13 and 8/13. No-propagation adopted it in 0/13 and listed the superseded value in 8/13 for both models; quarantine adopted it in 0/13 because invalidating without recomputing removes the line. On the \emph{success} endpoint (adopted and every other medication identity retained) re-reading scored 9/13 and 10/13, below its adoption count, because it omitted other lines more often than rebuild and local repair (which kept them in 12/13 per cell). The batch's pre-declared primary comparison (development evidence), local repair against re-reading on success, was 8 vs 9 (Qwen; discordant 2/3; $p=1.0$) and 7 vs 10 (Mistral; 2/5; $p=0.45$, Holm-2 $p=0.91$): no detected difference at $n=13$. Against the two controls (manipulation checks; neither adopted in any unit), local repair differed for Qwen (8 vs 0, Holm-6 $p=0.047$) and not for Mistral (7 vs 0, Holm-6 $p=0.0625$). In all 18 rebuild and local-repair units that did not adopt, the new value was in the facts shown, so it had propagated and was left out of the answer; 16 of the 18 (five patients per model) had also left that line out before the revision, an association rather than an explanation (App.~\ref{app:batchC}, which also notes a spelling variant scored as missing plus unsupported).
\paragraph{Shared-ingest comparison and an adapted selector (batch E).} A pre-specified batch forked rebuild, local repair and the adapted value--cost selector of App.~\ref{app:adapted} ($\lambda=0.3$ and $10$) from one shared ingested state per unit (13 replacement and 8 revocation units, 17 patients; 210 trajectories, none failed; App.~\ref{app:batchE}). The adoption gap persisted: re-reading 13/13 for both models; rebuild and local repair 8--9/13 (Qwen) and 10/13 (Mistral). Between batches C and E (separate ingests, same 13 patients) memory adoption moved by 1--2 per cell, as much as memory pipelines differed.

\paragraph{Pre-specified replacement follow-up (C2).} Both models were evaluated on 8 calibration patients not previously used by the agent protocol (A) and 11 reused held-out patients (B), with five pipelines and a frozen scorer (190 trajectories, no failures). Re-reading adopted every replacement, but Mistral A success was 4/8 versus 6/8 for rebuild and local repair. None of the pre-specified primary paired adoption tests rejected the null ($p=0.0625,0.5,0.0625,0.25$ for Qwen A, Mistral A, Qwen B and Mistral B); no secondary test survived Holm-3 (minimum adjusted $p=0.1875$). Figure~\ref{fig:c2} and Table~\ref{tab:batchCtwo} (App.~\ref{app:batchCtwo}) give adoption and success separately.
 The runner made three use calls where the frozen plan specified two; adoption and success, scored on the second, are unaffected, and C2 cost is not used (App.~\ref{app:batchCtwo}). An exploratory 32B repeat on reused records, run after the 7B results, rejected the C2 primary in two of three cells (per-cell $\alpha=0.05$; one after Holm-3; App.~\ref{app:largemodel}).

\subsection{Problem-list scoring: post-hoc lexical sensitivity analysis}\label{sec:measure}
\paragraph{Instruments.} Both earlier scorers accepted wordings that a record-fidelity reading rejects, so we fixed a versioned record-fidelity contract (a path counts only when its entity and every recorded qualifier appear) and three amendments, each written after outputs had been seen (App.~\ref{app:contract}).

\paragraph{Re-scoring (post hoc).} Table~\ref{tab:rules} (App.~\ref{app:rescoring}) is a scorer-sensitivity analysis of E1 under seven lexical rules, not confirmatory evidence of record quality. Of 12 per-model memory-versus-re-reading comparisons (Holm-12), \vLexSigRflenientQwen{} to \vLexSigProxyfrozenMistral{} were significant depending on the rule (Tables~\ref{tab:rulesfull} and~\ref{tab:rulesfullB}); re-reading was at or above the memory pipelines in every cell, but the counts are properties of the rule. Every rule is lexical; none is a pre-specified semantic endpoint or clinical adjudication.

\paragraph{Same-facts ablation (batch A) and decoding replay (batch B).} Batch A cloned one facts-and-summaries snapshot per patient, condition and model into four use branches (facts only, F; summaries then facts, SF; facts then summaries, FS; current documents, D). No patient-paired SF$-$F difference in strict record fidelity was significant under any rule, and Mistral's facts-only branch F still lay below D (8 vs 12 and 8 vs 11 of 16; Holm $p=1.0$). Batch B re-issued the use calls at temperature 0.2 with two seeds: 1--2 of 128 fidelity-strict outcomes per model flipped and no primary cell was significant (App.~\ref{app:batchAB}).

\section{Discussion}\label{sec:discussion}
\paragraph{What this comparison shows.} On short records local repair saved tokens against rebuilding but not against re-reading (\S\ref{sec:cost}). At about 10{,}000 tokens memory's per-use term turned negative, but the saving came from replacing documents the task cannot use by short, partly truncated extractions, not from compressing the evidence it uses: task-filtered memory still cost 1.4--1.5$\times$ source-filtered re-reading per use. The baseline to beat is therefore the strongest re-reading pipeline (source filtering, cached input), not re-reading everything at full price. On correctness, re-reading's 7B adoption lead held in direction in every C2 cell but reached $\alpha=0.05$ in none, came without a significant lead on success, and in batches C and E every rebuild or repair miss left out a value its prompt contained, which locates the miss at use, not in propagation, but does not explain it.
\paragraph{Alternative explanations not excluded.} The extractor types every drug mention as a medication fact whatever its parent's source (a treatment-record drug becomes \texttt{med\,|\,\ldots{}\,|\,from=D4:ehr}), and problem-list facts appear as category-prefixed paths, so the E4 additions and the problem-list gap may be effects of extraction typing and presentation rather than of maintenance: hypotheses for an ablation, not findings.
\section{Limitations}\label{sec:limits}
One dataset; two 7B models, with one later 32B exploratory supplement (App.~\ref{app:largemodel}); greedy build stage; 3--20 development, calibration or sealed patients per cell, reused across pipelines and conditions. The length sweep reused the eight screening patients (3--8 per level, each level on its own patients), added only non-allowed documents, and weighted input and output tokens equally; its analysis program omitted one pre-specified prediction, and its cached-input sensitivity is valid for Qwen only (App.~\ref{app:length}). Scorers are lexical and were written after inspecting development outputs; no clinical adjudication exists; all contract and amendment numbers are post-hoc re-scorings of seen outputs. Replacement uses counterfactual strength changes only; there is no adversarial evaluation, one revision per trajectory, one repeated use template, and costs are tokens, not prices or latency. Greedy decoding was not bit-reproducible on the serving stack. Persistence, extraction/typing, summary, source rendering and prompts are not independently manipulated, so no causal maintenance effect is identified.
\section{Conclusion}\label{sec:conclusion}
On short records re-reading had the lower full-pipeline token cost; with about 10{,}000 tokens of documents the task does not use, memory's mean cumulative cost undercut full-record re-reading after 2--14 uses, partly through truncated extraction, but never source-filtered re-reading, which by design skipped those documents. Re-reading adopted replacements more often, without a significant 7B lead on success; neither confirmatory analysis (the held-out E4 primary tests and the four C2 primary tests) rejected, and since representations, prompts and persistence differ together, none of this shows that repair itself is ineffective. Progress still requires one pre-specified semantic endpoint, larger unseen samples, records that grow in allowed evidence and an information-matched compressive ablation.
\subsection*{AI use statement}
Following the ICLR 2027 AI policy, each listed task is marked used, not used or not applicable. The language models evaluated in the experiments are the object of study and are not counted here. \emph{Required disclosures.} Generate synthetic data sets: not used; evaluation scenarios, including the counterfactual revision events, are deterministic rule-based transformations of public de-identified records, produced by AI-assisted code. Help develop theoretical models or conceptual frameworks: used, to refine the evaluation framing and the cost-accounting decomposition. Formulate mathematical claims: used, to state and check the accounting identity $\Delta C(k)$ of \S\ref{sec:cost}. Provide critical ingredients for proving mathematical claims and assist in the writing of proofs: not applicable; the paper proves no theorem. Propose or refine hypotheses: used, including drafting the pre-specified predictions of the length sweep, which the author fixed before any model call. Design or provide feedback on research methodology or experiments: used. Implement methods: used, for the evaluation harness, scenario builder and analysis programs, which were checked against unit tests and pre-declared controls. Assist with translation: used. Clean and reformat dataset: used, through AI-assisted code that extracts and renders the eICU tables; no record content was edited by hand or by a model. Support qualitative and thematic data analysis: used only for model-assisted audits of outputs, treated as internal checks; every reported endpoint comes from a deterministic scorer. Interpret results: used. \emph{Recommended disclosures.} Formulate questions for surveys or interviews, and transcribe recordings of research material: not applicable. Create or modify scientific figures or images, suggest experimental parameters, create or edit software code, and create artefacts: used. Draft parts of the paper, edit it for readability and suggest its structure: used; every statement was checked against the recorded results. Propose or refine the title or keywords: not used. Summarise or analyse existing literature, identify relevant literature, brainstorm, search for information, and format references: used; every reference was checked against its source. Discover research topics or identify gaps: not used. Reported values were recomputed from recorded outputs. The author reviewed all AI-assisted work and takes responsibility for the final text, claims, analyses, citations and artefacts.

\subsection*{Ethics statement}
The records are de-identified, publicly released ICU data (eICU Collaborative Research Database demo v2.0.1, Open Data Commons Open Database License v1.0; no credentialing is required by the data provider); no new patient data were collected and no interaction with patients or clinicians took place, so no ethics approval was sought. All revision events are rule-based counterfactual edits of records, and every clinical-sounding output in this study is an evaluation artefact, not advice: we claim no clinical validity and no endpoint is a clinical adjudication. Any model outputs or scenario renderings shared during review or released later are derived from the demo data and released under its terms; the eICU event tables are not redistributed.

\subsection*{Reproducibility statement}
Each batch used frozen code and scenarios and a scoring contract hashed before output, and recorded raw trajectories containing prompts, responses, usage, snapshots and read sets. These records reconstruct every held-out cell and its labels; re-scoring needs the scorer code. The length sweep's plan, build, scenario manifest and analysis program were hashed before its first model call (App.~\ref{app:length}). The paper specifies the split, events, pipelines, models, decoding, endpoints and tests. A fresh model run needs the code, public weights, vLLM, a GPU and the open-access eICU demo tables, which we do not redistribute. No code or data accompany this preprint. Code will be released upon acceptance, together with the archive described in App.~\ref{app:artifact} (scenarios, raw trajectories, per-example outcomes and generated tables).

\bibliographystyle{iclr2027_conference}
\bibliography{refs}

\newpage
\appendix
\renewcommand{\topfraction}{0.9}\renewcommand{\bottomfraction}{0.8}\renewcommand{\textfraction}{0.07}\renewcommand{\floatpagefraction}{0.8}\setcounter{topnumber}{4}\setcounter{bottomnumber}{3}\setcounter{totalnumber}{6}\raggedbottom
\makeatletter\setlength{\@fptop}{0pt}\setlength{\@fpsep}{14pt plus 1fil}\setlength{\@fpbot}{0pt plus 1fil}\makeatother
\section{Harness details and tested properties}\label{app:harness}\suppressfloats[t]
\paragraph{Store.} Evidence rows carry (id, version, source, patient, payload digest, status); node rows carry (id, version, type, content digest, parents with the parent versions seen, status, creating session, snapshot id). A snapshot is one read transaction over active evidence and active nodes; its identifier is a content hash. Claims are admitted through a gateway (authorised source, i.e.\ a connector source the harness may ingest, distinct from a task's allowed sources; same patient namespace; active evidence) inside one immediate transaction that rejects the whole batch as \emph{stale input} if any shown evidence changed version. The commit gate validates every evidence version and every node's own parent versions from the snapshot read set inside one transaction with the action record. It ensures only that no row in this read set changed between the snapshot and the commit; it does not check support, source permission, factual correctness or whether a rendered node reflects the latest document version (under no-propagation, nodes derived from a superseded version stay renderable and their actions commit), and it never blocked a commit in the sequential runs reported here.
\paragraph{Six deterministic properties (scripted clients, no model).} H1 revision after snapshot blocks the commit; H2 an unrelated write does not; H3 stale-input rejection of extraction; H4 exact transitive-successor invalidation; H5 dependency-ordered local rebuild; H6 read-set constraints are per node, never flattened. A seventh scripted case demonstrates a limit: unsupported content emitted by an extractor carries its chunk's parent and survives another document's revocation.
\paragraph{Prompts.} Extraction: one document per call, the document shown with its connector source, patient and version tag, returning candidate facts with an evidence reference; references outside the shown document are gated out. Summaries: per document from the admitted facts, then a global summary from these per-document (chunk) summaries. Use: the memory pipelines show the summaries and the facts as \texttt{- <type> | <content> | claim=<id> | from=<document>:<source>}; the version of each document is tracked by the host read set and is not shown to the model. Re-reading shows the current documents with their source and version tags. The allowed-source sentence is identical across pipelines; the task sentence differs only by the phrase ``from these documents'' appended for re-reading; all pipelines share one system prompt, which was written for the memory setting (``You have persistent memory about ONE patient from earlier sessions, shown below'') and is therefore not a perfect description of the re-reading prompt. Every extracted drug mention is typed as a medication fact whatever its parent's connector source, so a drug named in a treatment record appears in the memory prompt as a medication fact tagged \texttt{from=D4:ehr}.
\paragraph{Parse repair.}\label{app:parse} Of the extraction calls of the screening jobs, most parsed strictly; Mistral's outputs frequently lacked a closing brace and were repaired by a labelled deterministic parser that never alters a kept field; length-capped outputs keep only a complete prefix, so parseable is not complete. Every repair is recorded per call. In the replacement batch Mistral used the repair on 130 of 273 extractions and Qwen on 0.

\FloatBarrier
\section{Artefact availability and event-level audit trail}\label{app:artifact}\suppressfloats[t]
The evidence archive is a hash-addressed archive with a manifest and checksum ledger. It packages the frozen E1/E1b/E2/E4 scenario inputs (pseudonymous patient identifiers, rendered evidence documents, active versions, revision event, target support class and expected endpoint), the raw run directories (prompts, responses, per-call usage, snapshots, read sets, parent versions, commit records and run status), the outcome labels and scorer explanations, and the pre-specification files and amendments; it does not contain source code. Any copy provided to reviewers passes identifying strings through a consistent anonymisation map, and its ledger is authoritative for that copy. Neither the archive nor the code is public at the time of this preprint; both will be released upon acceptance (App.~\ref{app:repro}).

With our audit scripts (not in the archive), from these raw records the reported held-out medication cells and E4 primary tests are re-derived, the raw model text re-scored, E4 prompt provenance verified, the C2 scorer checked against destructive corruptions, and the reuse and 32B summaries regenerated; the frozen scenario files are sufficient to audit the revision labels and denominators. Without source code, this supports reconstruction of the reported results from the raw records but not a fresh model run, which also needs public weights, vLLM, a GPU and the open-access eICU demo tables (not redistributed). The event-level records, raw trajectories and outcome labels are the auditable evidence.

\FloatBarrier
\section{Held-out run: pre-specification and full tables (with development retention counts)}\label{app:heldout}\suppressfloats[t]
\paragraph{Development E1b retention counts.} The still-supported drug was retained in 5--8 of 8 episodes in every E1b cell (small size: Qwen 7 of 8 in every pipeline, Mistral 7 for re-read and 6 for each memory pipeline; large size: Qwen rebuild and local repair 8, re-read and cache 6; Mistral re-read 7, cache 6, rebuild and local repair 5).

The pre-specification fixed, before any sealed patient was touched by the agent protocol (section G of the pre-specification file records that the sealed partition's records had been used in an earlier project split): the questions; the four pipelines; the four conditions at the small size; one decoding; the frozen code build; the primary endpoint (E4 unsupported-source addition per patient, each memory pipeline against re-reading, one-sided exact McNemar with Holm correction over the three memory pipelines within a model); the sample (all eligible sealed patients, amended once before generation to the 20 eligible); and the stop rules. Scenarios were generated once; the two jobs were submitted once and read once. An independent recomputation found 0 discrepancies in 447 checks. \emph{Erratum to the pre-specification text (recorded 2026-09-17, after the run; the original file and its hash are kept unchanged).} Its sample-size rationale stated that the one-sided exact McNemar test rejects at $\alpha=0.05$ with $\ge5$ discordant pairs all in the hypothesised direction ``or $\ge7$ of 8, $\ge8$ of 10''. Exact values: 5:0 gives $p=0.03125$, 7:1 gives $p=0.0352$, but 8:2 gives $p=0.0547$, which does not reject; and under the pre-declared Holm correction over three pipelines the first-step threshold is $0.05/3\approx0.0167$, which 5:0 does not pass (6:0, $p=0.0156$, is the smallest all-one-direction pattern that does). The test actually applied is the pre-declared Holm-corrected one; every observed $p$ was $\ge0.25$, so no conclusion changes, and the rationale was a description of power, not of the test. The pre-specification is a hashed file in the project record; no externally time-stamped registration exists, so the ``fixed before the run'' statement rests on the project's own file records. Table~\ref{tab:heldoutprimary} gives the pre-specified primary test and Table~\ref{tab:heldoutfull} every held-out cell.

\begin{table}[!tbp]\centering
\caption{Pre-specified held-out E4 primary test: each memory pipeline is compared with re-reading on patient-level unsupported-source additions (20 sealed patients per model). The table reports the memory/re-read counts, discordant pairs (m./r.\ = memory-only/re-read-only), one-sided exact McNemar $p$ values and Holm-adjusted thresholds (thr.) for the three within-model comparisons. Non-rejection is not evidence of equivalence.}\label{tab:heldoutprimary}
\footnotesize\setlength{\tabcolsep}{4pt}\begin{tabular}{llcccccl}\toprule
model & memory pipeline & memory & re-read & discordant m./r. & $p$ (one-sided) & Holm thr. & rejected\\\midrule
Qwen & cache & 2/20 & 0/20 & 2 / 0 & 0.25 & 0.0167 & no\\
Qwen & rebuild & 0/20 & 0/20 & 0 / 0 & 1.0 & 0.05 & no\\
Qwen & local repair & 1/20 & 0/20 & 1 / 0 & 0.5 & 0.025 & no\\
Mistral & cache & 1/20 & 4/20 & 1 / 4 & 0.9688 & 0.05 & no\\
Mistral & rebuild & 4/20 & 4/20 & 3 / 3 & 0.6562 & 0.0167 & no\\
Mistral & local repair & 4/20 & 4/20 & 3 / 3 & 0.6562 & 0.025 & no\\
\bottomrule\end{tabular}

\end{table}
\begin{table}[!tbp]\centering
\caption{Held-out run, full table: 20 sealed patients, small size, four conditions, four pipelines, two models; medication use after the revision point (E0 and E4 have no revision). config = pipeline; strict = strict completion; counts of 20 with exact 95\,\% Clopper--Pearson intervals for the proportion; stale is defined only for E1 and retained only for E1b (--- = not defined in that condition); ret.\ = retained, i.e.\ still-supported key present after the E1b revocation; unsup., fabr.\ and miss.\ = patients whose list named a drug found only in non-allowed documents, named a drug found in no document, or lacked an expected key, respectively; dx = problem-list (diagnosis) use, scored here by the frozen leaf-match proxy; tokens = mean prompt+completion per trajectory.}\label{tab:heldoutfull}
\footnotesize\setlength{\tabcolsep}{3pt}\begin{tabular}{llccccccccr}\toprule
model & cond. & config & strict [CI] & stale & unsup. [CI] & fabr. & miss. & ret. & dx & tokens\\\midrule
Qwen & E0 & re-read & 18 [0.68--0.99] & --- & 0 [0.00--0.17] & 2 & 2 & --- & 18 & 1764\\
Qwen & E0 & cache & 15 [0.51--0.91] & --- & 0 [0.00--0.17] & 3 & 5 & --- & 15 & 3797\\
Qwen & E0 & rebuild & 17 [0.62--0.97] & --- & 0 [0.00--0.17] & 2 & 3 & --- & 9 & 6295\\
Qwen & E0 & local repair & 19 [0.75--1.00] & --- & 0 [0.00--0.17] & 1 & 1 & --- & 9 & 6250\\
Qwen & E1 & re-read & 18 [0.68--0.99] & 0 & 0 [0.00--0.17] & 2 & 2 & --- & 18 & 1827\\
Qwen & E1 & cache & 15 [0.51--0.91] & 0 & 0 [0.00--0.17] & 2 & 5 & --- & 14 & 4129\\
Qwen & E1 & rebuild & 18 [0.68--0.99] & 0 & 0 [0.00--0.17] & 1 & 2 & --- & 9 & 9717\\
Qwen & E1 & local repair & 17 [0.62--0.97] & 0 & 0 [0.00--0.17] & 2 & 3 & --- & 9 & 7375\\
Qwen & E1b & re-read & 18 [0.68--0.99] & --- & 0 [0.00--0.17] & 2 & 2 & 18/20 & 18 & 1805\\
Qwen & E1b & cache & 15 [0.51--0.91] & --- & 0 [0.00--0.17] & 3 & 5 & 18/20 & 15 & 4080\\
Qwen & E1b & rebuild & 18 [0.68--0.99] & --- & 0 [0.00--0.17] & 1 & 2 & 19/20 & 9 & 9663\\
Qwen & E1b & local repair & 17 [0.62--0.97] & --- & 0 [0.00--0.17] & 2 & 3 & 18/20 & 9 & 7311\\
Qwen & E4 & re-read & 18 [0.68--0.99] & --- & 0 [0.00--0.17] & 2 & 2 & --- & 18 & 1911\\
Qwen & E4 & cache & 16 [0.56--0.94] & --- & 2 [0.01--0.32] & 2 & 3 & --- & 15 & 4173\\
Qwen & E4 & rebuild & 18 [0.68--0.99] & --- & 0 [0.00--0.17] & 2 & 2 & --- & 11 & 6958\\
Qwen & E4 & local repair & 17 [0.62--0.97] & --- & 1 [0.00--0.25] & 2 & 3 & --- & 9 & 6971\\
\midrule
Mistral & E0 & re-read & 14 [0.46--0.88] & --- & 0 [0.00--0.17] & 3 & 6 & --- & 15 & 1945\\
Mistral & E0 & cache & 13 [0.41--0.85] & --- & 0 [0.00--0.17] & 4 & 7 & --- & 7 & 4551\\
Mistral & E0 & rebuild & 16 [0.56--0.94] & --- & 0 [0.00--0.17] & 2 & 4 & --- & 3 & 7390\\
Mistral & E0 & local repair & 16 [0.56--0.94] & --- & 0 [0.00--0.17] & 2 & 4 & --- & 2 & 7398\\
Mistral & E1 & re-read & 14 [0.46--0.88] & 0 & 0 [0.00--0.17] & 3 & 6 & --- & 15 & 2018\\
Mistral & E1 & cache & 13 [0.41--0.85] & 0 & 0 [0.00--0.17] & 4 & 7 & --- & 7 & 4924\\
Mistral & E1 & rebuild & 16 [0.56--0.94] & 0 & 0 [0.00--0.17] & 2 & 4 & --- & 2 & 11607\\
Mistral & E1 & local repair & 16 [0.56--0.94] & 0 & 0 [0.00--0.17] & 2 & 4 & --- & 3 & 8676\\
Mistral & E1b & re-read & 14 [0.46--0.88] & --- & 0 [0.00--0.17] & 3 & 6 & 19/20 & 15 & 1996\\
Mistral & E1b & cache & 13 [0.41--0.85] & --- & 0 [0.00--0.17] & 4 & 7 & 18/20 & 7 & 4888\\
Mistral & E1b & rebuild & 16 [0.56--0.94] & --- & 0 [0.00--0.17] & 2 & 4 & 18/20 & 3 & 11538\\
Mistral & E1b & local repair & 16 [0.56--0.94] & --- & 0 [0.00--0.17] & 2 & 4 & 18/20 & 2 & 8609\\
Mistral & E4 & re-read & 11 [0.32--0.77] & --- & 4 [0.06--0.44] & 3 & 5 & --- & 16 & 2122\\
Mistral & E4 & cache & 14 [0.46--0.88] & --- & 1 [0.00--0.25] & 3 & 6 & --- & 8 & 4998\\
Mistral & E4 & rebuild & 14 [0.46--0.88] & --- & 4 [0.06--0.44] & 2 & 2 & --- & 2 & 8226\\
Mistral & E4 & local repair & 12 [0.36--0.81] & --- & 4 [0.06--0.44] & 4 & 4 & --- & 2 & 8243\\
\bottomrule\end{tabular}

\end{table}
\paragraph{Per-phase accounting.} Table~\ref{tab:heldoutphases} splits every held-out cell's mean tokens by phase from the raw call usage. The 3.5--4.3$\times$ of \S\ref{sec:cost} is the full-trajectory ratio (ingest, revision and the three use calls); the use-stage prompt tokens alone are 2.1--2.2$\times$ re-reading's (2.09--2.23 over the eight model--condition cells), and use-stage prompt+completion 1.9--2.0$\times$; ingest and revision account for the rest.
\begin{table}[!tbp]\centering
\caption{Held-out run, per-phase mean prompt+completion tokens per trajectory (20 sealed patients per cell; raw \texttt{llm\_call} usage). use prompt / use compl.\ = prompt and completion tokens of the three use calls; the last two columns are the ratios to re-reading of the total and of the use-stage prompt tokens. Phase means are rounded separately, so phases can differ from the total by one token; ratios use unrounded means.}\label{tab:heldoutphases}
\fontsize{8}{9.5}\selectfont\setlength{\tabcolsep}{3pt}\begin{tabular}{@{}lllrrrrrrr@{}}\toprule
model & cond. & pipeline & ingest & revise & use prompt & use compl. & total & $\times$ re-read (total) & $\times$ re-read (use prompt)\\\midrule
Qwen & E0 & re-read & 0 & 0 & 1466 & 298 & 1764 & 1.00 & 1.00\\
Qwen & E0 & cache & 1540 & 0 & 1969 & 289 & 3797 & 2.15 & 1.34\\
Qwen & E0 & rebuild & 2862 & 0 & 3146 & 286 & 6295 & 3.57 & 2.15\\
Qwen & E0 & local repair & 2850 & 0 & 3118 & 283 & 6250 & 3.54 & 2.13\\
\midrule
Qwen & E1 & re-read & 0 & 0 & 1517 & 310 & 1827 & 1.00 & 1.00\\
Qwen & E1 & cache & 1811 & 0 & 2017 & 301 & 4129 & 2.26 & 1.33\\
Qwen & E1 & rebuild & 3351 & 2856 & 3226 & 285 & 9717 & 5.32 & 2.13\\
Qwen & E1 & local repair & 3347 & 509 & 3234 & 285 & 7375 & 4.04 & 2.13\\
\midrule
Qwen & E1b & re-read & 0 & 0 & 1511 & 294 & 1805 & 1.00 & 1.00\\
Qwen & E1b & cache & 1789 & 0 & 2003 & 288 & 4080 & 2.26 & 1.33\\
Qwen & E1b & rebuild & 3307 & 2861 & 3211 & 284 & 9663 & 5.35 & 2.13\\
Qwen & E1b & local repair & 3308 & 507 & 3211 & 286 & 7311 & 4.05 & 2.13\\
\midrule
Qwen & E4 & re-read & 0 & 0 & 1611 & 300 & 1911 & 1.00 & 1.00\\
Qwen & E4 & cache & 1800 & 0 & 2070 & 303 & 4173 & 2.18 & 1.29\\
Qwen & E4 & rebuild & 3319 & 0 & 3351 & 287 & 6958 & 3.64 & 2.08\\
Qwen & E4 & local repair & 3323 & 0 & 3361 & 287 & 6971 & 3.65 & 2.09\\
\midrule
Mistral & E0 & re-read & 0 & 0 & 1607 & 339 & 1945 & 1.00 & 1.00\\
Mistral & E0 & cache & 2043 & 0 & 2161 & 347 & 4551 & 2.34 & 1.35\\
Mistral & E0 & rebuild & 3520 & 0 & 3577 & 293 & 7390 & 3.80 & 2.23\\
Mistral & E0 & local repair & 3523 & 0 & 3581 & 294 & 7398 & 3.80 & 2.23\\
\midrule
Mistral & E1 & re-read & 0 & 0 & 1664 & 355 & 2018 & 1.00 & 1.00\\
Mistral & E1 & cache & 2365 & 0 & 2209 & 350 & 4924 & 2.44 & 1.33\\
Mistral & E1 & rebuild & 4100 & 3523 & 3687 & 297 & 11607 & 5.75 & 2.22\\
Mistral & E1 & local repair & 4098 & 597 & 3686 & 296 & 8676 & 4.30 & 2.22\\
\midrule
Mistral & E1b & re-read & 0 & 0 & 1658 & 338 & 1996 & 1.00 & 1.00\\
Mistral & E1b & cache & 2348 & 0 & 2205 & 335 & 4888 & 2.45 & 1.33\\
Mistral & E1b & rebuild & 4060 & 3523 & 3664 & 290 & 11538 & 5.78 & 2.21\\
Mistral & E1b & local repair & 4059 & 597 & 3664 & 289 & 8609 & 4.31 & 2.21\\
\midrule
Mistral & E4 & re-read & 0 & 0 & 1775 & 346 & 2122 & 1.00 & 1.00\\
Mistral & E4 & cache & 2359 & 0 & 2285 & 353 & 4998 & 2.36 & 1.29\\
Mistral & E4 & rebuild & 4074 & 0 & 3861 & 292 & 8226 & 3.88 & 2.18\\
Mistral & E4 & local repair & 4076 & 0 & 3872 & 295 & 8243 & 3.89 & 2.18\\
\bottomrule
\end{tabular}

\end{table}

\FloatBarrier
\section{Detailed E1 phase costs}\label{app:costdetail}\suppressfloats[t]
\begin{figure}[!tbp]\centering
\includegraphics[width=\textwidth]{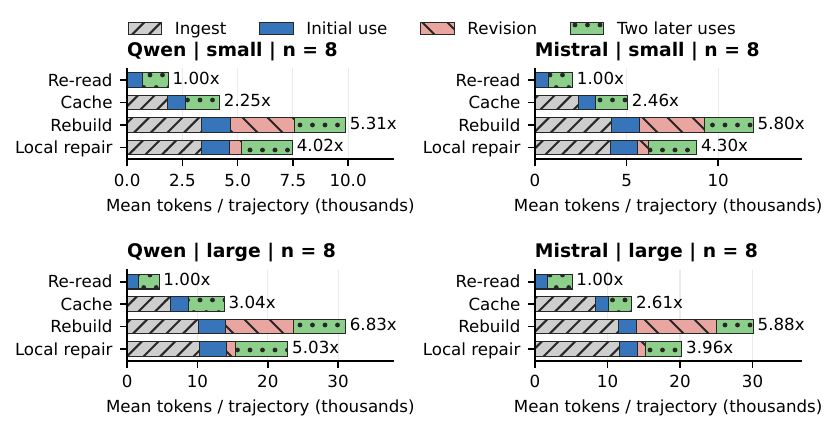}
\caption{E1 development token costs, recomputed from raw call usage: eight patients per cell, shared across settings. Bars separate ingestion, initial use, revision and the two later uses; labels show total cost relative to re-reading within each panel. These are descriptive means, not independent replications or projected reuse curves. Exact values and call counts: Table~\ref{tab:cost}.}\label{fig:cost}
\end{figure}
Table~\ref{tab:cost} gives the per-phase token and model-call breakdown of the E1 development cells summarised in \S\ref{sec:cost}.
\begin{table}[!tbp]\centering
\caption{E1 development cells: mean prompt+completion tokens per trajectory by phase (use$_0$ = initial use before the revision; use$_1$, use$_2$ = the two later uses) and model calls; inv./rebuilt = derived nodes invalidated / recomputed at the revision (Fig.~\ref{fig:pipeline}). Re-reading performs no ingest and no revision work. Phase means are rounded separately, so a row's phases can differ from its total by one token.}\label{tab:cost}
\footnotesize\setlength{\tabcolsep}{3pt}\begin{tabular}{@{}llrrrrrrr@{}}\toprule
setting & pipeline & ingest & use$_0$ & revise & use$_1$+use$_2$ & total & calls & inv./rebuilt\\\midrule
Qwen large & re-read & 0 & 1578 & 0 & 2970 & 4548 & 3 & ---\\
Qwen large & cache & 6159 & 2593 & 0 & 5060 & 13812 & 10.8 & 1/0\\
Qwen large & rebuild & 10152 & 3772 & 9721 & 7404 & 31050 & 34 & 3/14.5\\
Qwen large & local repair & 10253 & 3857 & 1296 & 7453 & 22858 & 20.5 & 3/1\\
\midrule
Mistral large & re-read & 0 & 1772 & 0 & 3345 & 5117 & 3 & ---\\
Mistral large & cache & 8402 & 1699 & 0 & 3230 & 13331 & 10.8 & 1/0\\
Mistral large & rebuild & 11509 & 2492 & 11043 & 5057 & 30101 & 34 & 3/14.5\\
Mistral large & local repair & 11591 & 2544 & 1116 & 4999 & 20250 & 20.5 & 3/1\\
\midrule
Qwen small & re-read & 0 & 684 & 0 & 1177 & 1861 & 3 & ---\\
Qwen small & cache & 1813 & 844 & 0 & 1526 & 4184 & 7 & 1/0\\
Qwen small & rebuild & 3379 & 1293 & 2896 & 2315 & 9882 & 19 & 3/7\\
Qwen small & local repair & 3367 & 1286 & 515 & 2315 & 7483 & 13 & 3/1\\
\midrule
Mistral small & re-read & 0 & 762 & 0 & 1292 & 2054 & 3 & ---\\
Mistral small & cache & 2365 & 963 & 0 & 1730 & 5057 & 7 & 1/0\\
Mistral small & rebuild & 4172 & 1528 & 3564 & 2654 & 11918 & 19 & 3/7\\
Mistral small & local repair & 4116 & 1489 & 602 & 2617 & 8825 & 13 & 3/1\\
\bottomrule\end{tabular}

\end{table}

\FloatBarrier
\section{Representation ratio}\label{app:rho}\suppressfloats[t]
The task-scoped representation ratio $\rho_{\text{task}}$ has a median above one in all 30 groups of the current screening runs (smallest 1.23). This is a CPU-filtered block ratio, not a task-filtered model run. The unscoped ratio $\rho_{\text{all}}$ falls below one in two Mistral large-cache groups (0.93 and 0.94), with extraction truncation and missing laboratory content. Neither ratio is the full per-use token cost (\S\ref{sec:cost}).

Two ratios are defined for a memory-pipeline use prompt and the re-reading prompt of the same scenario and use index, both counted with the model's own tokenizer on the prompt bodies (instruction text excluded): $\rho_{\text{all}}$ = memory-block tokens / tokens of \emph{all} active document bodies shown to re-reading; $\rho_{\text{task}}$ = tokens of the task-filtered memory block (medication facts and summaries of pharmacy documents for the medication task; problem-list facts and summary for the problem-list task) / tokens of the task-allowed document bodies. The task filter is a CPU reconstruction of the recorded blocks. Two measurements exist and are reported separately. (i) An earlier measurement (eight development patients; 144 pairs per job; 18 groups = job $\times$ task $\times$ memory pipeline): group medians $\ge 1$ for both ratios (smallest 1.20 for $\rho_{\text{all}}$, 1.01 for $\rho_{\text{task}}$); pairs below one: 6 of 432 ($\rho_{\text{all}}$) and 69 of 432 ($\rho_{\text{task}}$). (ii) The screening batches of this paper (768 recorded use prompts, 576 pairs, 30 groups; tokenised in a CPU job whose chat-template token counts equal the served prompt tokens for 768 of 768 prompts): $\rho_{\text{task}}$ medians $\ge 1$ in all 30 groups (smallest 1.23; 6 of 576 pairs below one); $\rho_{\text{all}}$ medians $\ge 1$ in 28 groups, below one in the two Mistral large-size cache groups (0.93 and 0.94; 24 of 576 pairs below one, all in these groups), where 11 of the 116 cache extraction calls ended at the 2{,}048-token cap and the memory therefore lacks laboratory content---an incompleteness effect, not a compressive representation. Neither measurement supports a group-level compression advantage for maintained memory; the run directories of (i) are not in the App.~\ref{app:artifact} archive (their token counts are), those of (ii) are.

\FloatBarrier
\section{Problem-list scoring contract}\label{app:contract}\suppressfloats[t]
\paragraph{Instruments and their defects.} The screening runs scored the problem list with a leaf-match proxy (a truth path's last segment matched by an item), which accepts ``no chest pain'' and ``chest pain and sepsis'' for ``chest pain''. A record-fidelity scorer separated parent-only, bundled, negated and uncertain items, but a last-segment-only item (``acute'') passed its strict rule without naming the disease, a verbatim echo of a full path could be labelled bundled when sibling paths share a parent, and punctuation was removed before the bundle check; such a rule can itself create discordant pairs between pipelines.
\paragraph{Truth structure.} A path $s_1|s_2|c_1|\dots|c_k$ has a category ($s_1,s_2$; never sufficient) and a concept chain; the disease entity $E$ is the first chain segment that is not a qualifier (severity, laterality, cause, status such as ``s/p'', ``acute'', ``with \dots'', ranges); later chain segments are qualifiers, and every chain segment is task-required because the record states it.
\paragraph{Coverage and outcomes.} An item covers a path fully when it equals the path or its chain, or contains every chain segment as a whole phrase in any order (``acute hepatic dysfunction'' covers \texttt{hepatic dysfunction|acute}); it covers the entity when $E$ is present but a qualifier is missing (\emph{partial\_parent}, with a \emph{qualifier\_lost} flag); it is \emph{qualifier\_only} when only qualifiers are present. A full cover of one path is never a parent hit of a sibling path. Bundling is decided on fragments split at ``,'' ``;'' ``\&'' ``and'' before normalisation; a fragment inside the assigned path's own text is never an extra. Placeholders (\texttt{<\dots>}, ``\dots'', ``diagnosis value'') are \emph{uncertain}. Leading negation or status cues are retained when they are part of the covered path (``s/p CABG'') and otherwise make the item \emph{negated}. Outcome precedence: uncertain, negated, extra\_bundled, exact, partial\_parent, qualifier\_only, unmatched.
\paragraph{Endpoints.} \emph{strict}: format valid, committed, every path covered exactly, no other item. Relaxations recorded separately: $L_{\text{parent}}$ (partial\_parent covers), $L_{\text{extra}}$ (unmatched items allowed), $L_{\text{bundle}}$ (bundled covers); \emph{lenient} = all three; \emph{qualifier\_only}, \emph{negated} and \emph{uncertain} items never pass. Format failures, blocked and empty outputs are attempted, never success.
\paragraph{Summary and effect of the amendments.} Three amendments followed, each written after outputs had been seen and disclosed as such: amendment 1 (wording artefacts of our own rule), amendment 2 (exact-match priority for nested chains; postposed cues; parenthetical abbreviations only when traceable), and amendment 3 (cue scope per occurrence; constrained abbreviation matching), the last closing two synthetic false acceptances found by later reviews. Under amendment 2 a verbatim echo of all \vRuEchoRealListsN{} real truth lists passes (\vRuEchoRealListsPassSevenOne{} under amendment 1, \vRuEchoRealListsPassOld{} under fidelity); the amendment-2 defects changed no trajectory endpoint and one real item label (a batch-A ``DKA (Diabetic Ketoacidosis)'' item); amendment 3 changes \vRvStrictChanged{} endpoints and \vRvItemChanged{} item labels of the 1{,}618 recorded outputs.
\paragraph{Amendment 1.} Written after the contract re-scoring was read (disclosed): connectives are all function words; words of the path's recorded category are not extras; a parenthetical restatement or abbreviation of the covered path is not an extra; seven added controls (control set C15 of the scorer's test suite).
\paragraph{Amendment 2.} Written after the final independent audit (disclosed): an item equal to a full path or its chain is assigned to that path only, and nested chains are resolved by specificity (the more specific covered path wins when the item's content words lie within it); cue words are never connectives, and a postposed or mid-item cue (``pneumonia ruled out'', ``pneumonia; resolved'') makes the item negated unless the cue is part of a covered path segment; a parenthetical abbreviation is permitted only when traceable to the assigned path (initials of its words, or the token occurs in the path text), otherwise it is content and the item is bundled; capitalisation never grants permission; no external dictionary or model is used. Control sets C16--C19 include verbatim echoes of all 36 distinct real truth lists and the audit's ten adversarial cases. Results under every rule variant are reported.
\paragraph{Amendment 3.} Written after two later reviews (disclosed; post hoc like the earlier amendments): (a) a cue occurrence is retained only while the covered path segments contain at least as many occurrences of that cue phrase, so a second ``without'' that denies the entity makes the item negated while the recorded ``without hemodynamic compromise'' qualifier keeps its one occurrence; (b) a parenthesised token is a permitted abbreviation only if it occurs in the assigned path's text or its letters are word initials of the path's chain or category words in order, with at most one letter matched inside the previous word and only for tokens of three or more letters (COPD, CAD, DKA pass; PE and PUMA for ``pneumonia'' are extras). Ten added controls (control set C20); 111 controls pass; the re-scoring of all 1{,}618 recorded outputs changes \vRvStrictChanged{} endpoints and \vRvItemChanged{} item labels. A three-letter token with two initials and one in-word letter (``CAT'' for ``coronary artery disease'') is still accepted and recorded as a known residual of the lexical rule.
\paragraph{Controls.} Fourteen control families with success counterparts for every failure case (full-path echo of every truth list, chain-only echo, natural-order phrase, qualifier-only, entity without qualifier, four punctuation bundles, a legitimate comma, placeholders, documented status, negation, an extra disease, execution states, sibling paths, and the 16 batch-A truth lists: full-path echo must pass 16/16 and last-segment-only echo must fail wherever a qualifier leaf exists). The implementation was hashed after the controls passed and before any real output was re-scored.

\FloatBarrier
\section{Re-scoring under the contract and its amendments (post hoc)}\label{app:rescoring}\suppressfloats[t]
\paragraph{Where the facts were.} Of the 480 held-out memory-pipeline problem-list outputs, 289 fail a punctuation-normalised variant of the frozen proxy (299 fail the frozen proxy; \vRtMemFailStrict{} fail contract/amendment 1 strict; \vRuMemFailStrict{} fail amendment 2 and amendment 3 strict); in 265 of these 289 every reference leaf label was present as a normalised string in the facts shown to the model (24 partly, 0 none): the information was visible, and how the model uses it was not identified.

\begin{table}[!tbp]\centering
\caption{Post-hoc lexical scorer-sensitivity analysis of held-out problem-list outputs (\S\ref{sec:measure}), condition E1 (20 sealed patients per model). RR = re-read, C = cache, RB = rebuild, LR = local repair. Successful uses /20; $^{*}$ marks a memory pipeline separated from re-reading after Holm-12 correction over all four conditions within the model. These values are not a clinical or confirmatory quality endpoint. All conditions: Table~\ref{tab:rulesfull}.}\label{tab:rules}
\footnotesize\setlength{\tabcolsep}{4pt}\begin{tabular}{llccccccc}
\toprule
& & frozen & fidelity & fidelity & contract & amend.\,1 & amend.\,2 & amend.\,2\\
model & pipeline & proxy & strict & lenient & strict & strict & strict & lenient\\
\midrule
Qwen & RR & 18 & 12 & 19 & 16 & 16 & 18 & 19\\
Qwen & C & 14 & 10 & 19 & 13 & 13 & 15 & 19\\
Qwen & RB & 9$^{*}$ & 6 & 13 & 6$^{*}$ & 6$^{*}$ & 6$^{*}$ & 10\\
Qwen & LR & 9$^{*}$ & 6 & 12 & 6$^{*}$ & 6$^{*}$ & 6$^{*}$ & 9\\
\midrule
Mistral & RR & 15 & 11 & 19 & 13 & 13 & 15 & 19\\
Mistral & C & 7$^{*}$ & 4 & 12 & 5$^{*}$ & 5$^{*}$ & 6$^{*}$ & 11\\
Mistral & RB & 2$^{*}$ & 2$^{*}$ & 14 & 1$^{*}$ & 1$^{*}$ & 1$^{*}$ & 13\\
Mistral & LR & 3$^{*}$ & 2$^{*}$ & 14 & 2$^{*}$ & 2$^{*}$ & 2$^{*}$ & 14\\
\midrule
\multicolumn{2}{l}{Holm-12 sig.\ of 12: Qwen} & 7 & 0 & 0 & 8 & 8 & 8 & 0\\
\multicolumn{2}{l}{Holm-12 sig.\ of 12: Mistral} & 12 & 8 & 0 & 12 & 12 & 12 & 0\\
\bottomrule
\end{tabular}

\end{table}
\begin{table}[!tbp]\centering
\caption{Held-out problem-list outputs under seven scoring rules, all four conditions (20 sealed patients per model; post hoc), part A: frozen proxy, fidelity strict, fidelity lenient, contract strict. RR, C, RB, LR as in Table~\ref{tab:rules}; $^{*}$ = memory pipeline separated from re-reading after Holm-12 within the model; last row = number of significant comparisons per rule. Part B: Table~\ref{tab:rulesfullB}.}\label{tab:rulesfull}
\footnotesize\setlength{\tabcolsep}{2pt}\begin{tabular}{llcccccccccccccccc}
\toprule
& & \multicolumn{4}{c}{frozen proxy} & \multicolumn{4}{c}{fidelity strict} & \multicolumn{4}{c}{fidelity lenient} & \multicolumn{4}{c}{contract strict}\\
\cmidrule(lr){3-6}\cmidrule(lr){7-10}\cmidrule(lr){11-14}\cmidrule(lr){15-18}
model & cond. & RR & C & RB & LR & RR & C & RB & LR & RR & C & RB & LR & RR & C & RB & LR\\
\midrule
Qwen & E0 & 18 & 15 & 9$^{*}$ & 9$^{*}$ & 12 & 11 & 6 & 6 & 19 & 19 & 13 & 12 & 16 & 14 & 6$^{*}$ & 6$^{*}$\\
Qwen & E1 & 18 & 14 & 9$^{*}$ & 9$^{*}$ & 12 & 10 & 6 & 6 & 19 & 19 & 13 & 12 & 16 & 13 & 6$^{*}$ & 6$^{*}$\\
Qwen & E1b & 18 & 15 & 9$^{*}$ & 9$^{*}$ & 12 & 11 & 6 & 6 & 19 & 19 & 13 & 12 & 16 & 14 & 6$^{*}$ & 6$^{*}$\\
Qwen & E4 & 18 & 15 & 11 & 9$^{*}$ & 12 & 10 & 8 & 6 & 19 & 18 & 14 & 12 & 16 & 12 & 8$^{*}$ & 6$^{*}$\\
\midrule
Mistral & E0 & 15 & 7$^{*}$ & 3$^{*}$ & 2$^{*}$ & 11 & 4 & 2$^{*}$ & 1$^{*}$ & 19 & 12 & 13 & 14 & 13 & 5$^{*}$ & 1$^{*}$ & 1$^{*}$\\
Mistral & E1 & 15 & 7$^{*}$ & 2$^{*}$ & 3$^{*}$ & 11 & 4 & 2$^{*}$ & 2$^{*}$ & 19 & 12 & 14 & 14 & 13 & 5$^{*}$ & 1$^{*}$ & 2$^{*}$\\
Mistral & E1b & 15 & 7$^{*}$ & 3$^{*}$ & 2$^{*}$ & 11 & 4 & 2$^{*}$ & 1$^{*}$ & 19 & 12 & 14 & 13 & 13 & 5$^{*}$ & 1$^{*}$ & 1$^{*}$\\
Mistral & E4 & 16 & 8$^{*}$ & 2$^{*}$ & 2$^{*}$ & 12 & 6 & 1$^{*}$ & 1$^{*}$ & 18 & 13 & 12 & 10 & 14 & 6$^{*}$ & 1$^{*}$ & 1$^{*}$\\
\midrule
\multicolumn{2}{l}{Holm-12 sig.\ of 12} & \multicolumn{4}{c}{Qwen 7, Mistral 12} & \multicolumn{4}{c}{Qwen 0, Mistral 8} & \multicolumn{4}{c}{Qwen 0, Mistral 0} & \multicolumn{4}{c}{Qwen 8, Mistral 12}\\
\bottomrule
\end{tabular}

\end{table}
\begin{table}[!tbp]\centering
\caption{Part B of Table~\ref{tab:rulesfull}: amendment 1 strict, amendment 2 strict, amendment 2 lenient (same outputs, same conventions).}\label{tab:rulesfullB}
\footnotesize\setlength{\tabcolsep}{2.5pt}\begin{tabular}{llcccccccccccc}
\toprule
& & \multicolumn{4}{c}{amendment 1 strict} & \multicolumn{4}{c}{amendment 2 strict} & \multicolumn{4}{c}{amendment 2 lenient}\\
\cmidrule(lr){3-6}\cmidrule(lr){7-10}\cmidrule(lr){11-14}
model & cond. & RR & C & RB & LR & RR & C & RB & LR & RR & C & RB & LR\\
\midrule
Qwen & E0 & 16 & 14 & 6$^{*}$ & 6$^{*}$ & 18 & 16 & 6$^{*}$ & 6$^{*}$ & 19 & 19 & 10 & 9\\
Qwen & E1 & 16 & 13 & 6$^{*}$ & 6$^{*}$ & 18 & 15 & 6$^{*}$ & 6$^{*}$ & 19 & 19 & 10 & 9\\
Qwen & E1b & 16 & 14 & 6$^{*}$ & 6$^{*}$ & 18 & 16 & 6$^{*}$ & 6$^{*}$ & 19 & 19 & 10 & 9\\
Qwen & E4 & 16 & 12 & 8$^{*}$ & 6$^{*}$ & 18 & 14 & 9$^{*}$ & 7$^{*}$ & 19 & 17 & 11 & 9\\
\midrule
Mistral & E0 & 13 & 5$^{*}$ & 1$^{*}$ & 1$^{*}$ & 15 & 6$^{*}$ & 1$^{*}$ & 1$^{*}$ & 19 & 11 & 12 & 14\\
Mistral & E1 & 13 & 5$^{*}$ & 1$^{*}$ & 2$^{*}$ & 15 & 6$^{*}$ & 1$^{*}$ & 2$^{*}$ & 19 & 11 & 13 & 14\\
Mistral & E1b & 13 & 5$^{*}$ & 1$^{*}$ & 1$^{*}$ & 15 & 6$^{*}$ & 1$^{*}$ & 1$^{*}$ & 19 & 11 & 13 & 13\\
Mistral & E4 & 14 & 6$^{*}$ & 1$^{*}$ & 1$^{*}$ & 16 & 8$^{*}$ & 1$^{*}$ & 1$^{*}$ & 18 & 12 & 11 & 10\\
\midrule
\multicolumn{2}{l}{Holm-12 sig.\ of 12} & \multicolumn{4}{c}{Qwen 8, Mistral 12} & \multicolumn{4}{c}{Qwen 8, Mistral 12} & \multicolumn{4}{c}{Qwen 0, Mistral 0}\\
\bottomrule
\end{tabular}

\end{table}
\paragraph{Single relaxations (held-out, Holm-12 per model).} Under amendment 2 strict with exactly one relaxation, parent-level cover leaves \vRuSigParentQwen{} (Qwen) and \vRuSigParentMistral{} (Mistral) of the 12 comparisons significant, extra items \vRuSigExtraQwen{} and \vRuSigExtraMistral{}, bundled cover \vRuSigBundleQwen{} and \vRuSigBundleMistral{}.
\FloatBarrier
\paragraph{Causes of label changes (real outputs).} fidelity $\to$ contract: full-path echoes no longer labelled bundled (held-out / A / B: 49 / 25 / 50 outputs fail$\to$pass); last-segment-only items no longer passing (10 / 4 / 9 pass$\to$fail; one qualifier-only, the rest deeper-concept-only such as ``lactic acidosis'' for ``metabolic acidosis$|$lactic acidosis''); wording extras under the contract's literal fragment rule (3 / 8 / 16 pass$\to$fail); one further held-out partial-item flip (pass$\to$fail); punctuation bundles, placeholders and status cues: no real output. contract $\to$ amendment 1: 42 outputs change an item label (function words, restated categories, article inside a qualifier); strict flips: held-out 0, batch A 5 fail$\to$pass, batch B 8 fail$\to$pass, none pass$\to$fail; wording additions outside the recorded text (``\ldots including \ldots'', ``such as'') remain extras by the literal rule and account for the residual batch-A/B failures they cause (5 and 12). Uncertain items: 0 in 1,408 re-scored outputs.

\FloatBarrier
\section{Batch A and batch B details}\label{app:batchAB}\suppressfloats[t]
\paragraph{Batch A design.} One facts-and-summaries snapshot per patient, condition and model, built once with the real ingest path and cloned read-only into four use branches; the facts are byte-identical across F, SF and FS and the summaries across SF and FS (asserted at run time from recorded hashes). Each branch answers both tasks under E0 and E4 with one decoding on 16 development patients chosen by a pre-fixed eligibility rule and seed. The D branch shows the current documents with source and version tags; the memory branches show claim identifiers and connector sources, and versions are host-tracked. Integrity: 512 of 512 use outputs, 64 of 64 units passing the invariants, no format failure, blocked or empty output. No padding controlled the prompt-length difference between F and SF.
\paragraph{Counts behind \S\ref{sec:measure}.} On the patient-paired SF$-$F difference in strict record fidelity (four cells, one Holm family; Tables~\ref{tab:batchA} and~\ref{tab:batchAprimary}) no cell is significant under any rule, but the counts move with it: under fidelity strict, Qwen E0 11 vs 11 of 16, E4 10 vs 11, Mistral E0 6 vs 8, E4 7 vs 8 (Holm $p=1.0$ each); under amendment 2 strict, Qwen \vRuBatchASFQwenEZero{} vs \vRuBatchAFQwenEZero{} and \vRuBatchASFQwenEFour{} vs \vRuBatchAFQwenEFour{}, Mistral \vRuBatchASFMistralEZero{} vs \vRuBatchAFMistralEZero{} (discordant \vRuBatchADiscMistralEZero{}; Holm $p=\vRuBatchAHolmMistralEZero$) and \vRuBatchASFMistralEFour{} vs \vRuBatchAFMistralEFour{} (\vRuBatchADiscMistralEFour{}; $p=\vRuBatchAHolmMistralEFour$) of 16 (Table~\ref{tab:batchAamend2}). For Mistral the documents branch scored 12 and 11 of 16 (fidelity) and F lay below D under every rule, so the gap is present without the summary; whether the summary lowers fidelity further depends on how wording additions (``\ldots including \ldots'') are scored---they remain extras under amendment 2 and account for the residual SF failures---and we do not claim it.
\begin{table}[tbp]\centering
\caption{Batch A: diagnosis record fidelity (dx strict / lenient, fidelity scorer) and medication strict completion (med strict) per model, condition and branch (F, SF, FS, D as in \S\ref{sec:measure}).}\label{tab:batchA}
\footnotesize\setlength{\tabcolsep}{2.5pt}
\begin{tabular}{@{}lllrrrrrrrr@{}}\toprule
model & cond. & branch & scored & dx strict & dx lenient & med strict & fmt.\ fail & blocked & empty & missing/invalid\\\midrule
Qwen & E0 & F & 16+16 & 11/16 & 14/16 & 10/16 & 0 & 0 & 0 & 0/0\\
Qwen & E0 & SF & 16+16 & 11/16 & 15/16 & 13/16 & 0 & 0 & 0 & 0/0\\
Qwen & E0 & FS & 16+16 & 11/16 & 16/16 & 13/16 & 0 & 0 & 0 & 0/0\\
Qwen & E0 & D & 16+16 & 11/16 & 15/16 & 10/16 & 0 & 0 & 0 & 0/0\\
Qwen & E4 & F & 16+16 & 11/16 & 14/16 & 9/16 & 0 & 0 & 0 & 0/0\\
Qwen & E4 & SF & 16+16 & 10/16 & 15/16 & 14/16 & 0 & 0 & 0 & 0/0\\
Qwen & E4 & FS & 16+16 & 9/16 & 15/16 & 11/16 & 0 & 0 & 0 & 0/0\\
Qwen & E4 & D & 16+16 & 11/16 & 15/16 & 10/16 & 0 & 0 & 0 & 0/0\\
\midrule
Mistral & E0 & F & 16+16 & 8/16 & 14/16 & 10/16 & 0 & 0 & 0 & 0/0\\
Mistral & E0 & SF & 16+16 & 6/16 & 13/16 & 15/16 & 0 & 0 & 0 & 0/0\\
Mistral & E0 & FS & 16+16 & 6/16 & 12/16 & 11/16 & 0 & 0 & 0 & 0/0\\
Mistral & E0 & D & 16+16 & 12/16 & 16/16 & 11/16 & 0 & 0 & 0 & 0/0\\
Mistral & E4 & F & 16+16 & 8/16 & 14/16 & 11/16 & 0 & 0 & 0 & 0/0\\
Mistral & E4 & SF & 16+16 & 7/16 & 15/16 & 12/16 & 0 & 0 & 0 & 0/0\\
Mistral & E4 & FS & 16+16 & 5/16 & 12/16 & 11/16 & 0 & 0 & 0 & 0/0\\
Mistral & E4 & D & 16+16 & 11/16 & 16/16 & 13/16 & 0 & 0 & 0 & 0/0\\
\bottomrule\end{tabular}
\par\smallskip\parbox{\linewidth}{\footnotesize scored = valid diagnosis + medication outputs; fmt.\ fail, blocked, empty and missing/invalid are summed over both tasks; development evidence; exploratory.}

\end{table}
\begin{table}[!tbp]\centering
\caption{Batch A: primary and secondary patient-paired comparisons on diagnosis record fidelity (fidelity scorer) and, in the last family, on medication strict completion. A$-$B = the two branches compared; CP = exact Clopper--Pearson.}\label{tab:batchAprimary}
\fontsize{8}{9.5}\selectfont\setlength{\tabcolsep}{3pt}
\setlength{\tabcolsep}{3pt}\begin{tabular}{@{}llllllrrrr@{}}\toprule
model & cond. & A$-$B & metric & A: succ.\ (CP 95\,\%) & B: succ.\ (CP 95\,\%) & disc.\ A/B & diff & $p$ & Holm $p$\\\midrule
\multicolumn{10}{@{}l}{\emph{Primary family (Holm over 4): SF$-$F on diagnosis record fidelity, strict}}\\
Qwen & E0 & SF$-$F & dx strict & 11/16 (.41--.89) & 11/16 (.41--.89) & 1/1 & +0.00 & 1.000 & 1.000\\
Qwen & E4 & SF$-$F & dx strict & 10/16 (.35--.85) & 11/16 (.41--.89) & 1/2 & $-$0.06 & 1.000 & 1.000\\
Mistral & E0 & SF$-$F & dx strict & 6/16 (.15--.65) & 8/16 (.25--.75) & 0/2 & $-$0.12 & 0.500 & 1.000\\
Mistral & E4 & SF$-$F & dx strict & 7/16 (.20--.70) & 8/16 (.25--.75) & 1/2 & $-$0.06 & 1.000 & 1.000\\
\midrule
\multicolumn{10}{@{}l}{\emph{Secondary family, Qwen (Holm over 12): FS$-$SF, SF$-$D, F$-$D on strict and lenient}}\\
Qwen & E0 & FS$-$SF & dx strict & 11/16 (.41--.89) & 11/16 (.41--.89) & 2/2 & +0.00 & 1.000 & 1.000\\
Qwen & E4 & FS$-$SF & dx strict & 9/16 (.30--.80) & 10/16 (.35--.85) & 2/3 & $-$0.06 & 1.000 & 1.000\\
Qwen & E0 & SF$-$D & dx strict & 11/16 (.41--.89) & 11/16 (.41--.89) & 2/2 & +0.00 & 1.000 & 1.000\\
Qwen & E4 & SF$-$D & dx strict & 10/16 (.35--.85) & 11/16 (.41--.89) & 2/3 & $-$0.06 & 1.000 & 1.000\\
Qwen & E0 & F$-$D & dx strict & 11/16 (.41--.89) & 11/16 (.41--.89) & 1/1 & +0.00 & 1.000 & 1.000\\
Qwen & E4 & F$-$D & dx strict & 11/16 (.41--.89) & 11/16 (.41--.89) & 1/1 & +0.00 & 1.000 & 1.000\\
Qwen & E0 & FS$-$SF & dx lenient & 16/16 (.79--1.00) & 15/16 (.70--1.00) & 1/0 & +0.06 & 1.000 & 1.000\\
Qwen & E4 & FS$-$SF & dx lenient & 15/16 (.70--1.00) & 15/16 (.70--1.00) & 1/1 & +0.00 & 1.000 & 1.000\\
Qwen & E0 & SF$-$D & dx lenient & 15/16 (.70--1.00) & 15/16 (.70--1.00) & 1/1 & +0.00 & 1.000 & 1.000\\
Qwen & E4 & SF$-$D & dx lenient & 15/16 (.70--1.00) & 15/16 (.70--1.00) & 1/1 & +0.00 & 1.000 & 1.000\\
Qwen & E0 & F$-$D & dx lenient & 14/16 (.62--.98) & 15/16 (.70--1.00) & 0/1 & $-$0.06 & 1.000 & 1.000\\
Qwen & E4 & F$-$D & dx lenient & 14/16 (.62--.98) & 15/16 (.70--1.00) & 0/1 & $-$0.06 & 1.000 & 1.000\\
\midrule
\multicolumn{10}{@{}l}{\emph{Secondary family, Mistral (Holm over 12): FS$-$SF, SF$-$D, F$-$D on strict and lenient}}\\
Mistral & E0 & FS$-$SF & dx strict & 6/16 (.15--.65) & 6/16 (.15--.65) & 1/1 & +0.00 & 1.000 & 1.000\\
Mistral & E4 & FS$-$SF & dx strict & 5/16 (.11--.59) & 7/16 (.20--.70) & 1/3 & $-$0.12 & 0.625 & 1.000\\
Mistral & E0 & SF$-$D & dx strict & 6/16 (.15--.65) & 12/16 (.48--.93) & 0/6 & $-$0.38 & 0.031 & 0.375\\
Mistral & E4 & SF$-$D & dx strict & 7/16 (.20--.70) & 11/16 (.41--.89) & 0/4 & $-$0.25 & 0.125 & 1.000\\
Mistral & E0 & F$-$D & dx strict & 8/16 (.25--.75) & 12/16 (.48--.93) & 0/4 & $-$0.25 & 0.125 & 1.000\\
Mistral & E4 & F$-$D & dx strict & 8/16 (.25--.75) & 11/16 (.41--.89) & 0/3 & $-$0.19 & 0.250 & 1.000\\
Mistral & E0 & FS$-$SF & dx lenient & 12/16 (.48--.93) & 13/16 (.54--.96) & 1/2 & $-$0.06 & 1.000 & 1.000\\
Mistral & E4 & FS$-$SF & dx lenient & 12/16 (.48--.93) & 15/16 (.70--1.00) & 0/3 & $-$0.19 & 0.250 & 1.000\\
Mistral & E0 & SF$-$D & dx lenient & 13/16 (.54--.96) & 16/16 (.79--1.00) & 0/3 & $-$0.19 & 0.250 & 1.000\\
Mistral & E4 & SF$-$D & dx lenient & 15/16 (.70--1.00) & 16/16 (.79--1.00) & 0/1 & $-$0.06 & 1.000 & 1.000\\
Mistral & E0 & F$-$D & dx lenient & 14/16 (.62--.98) & 16/16 (.79--1.00) & 0/2 & $-$0.12 & 0.500 & 1.000\\
Mistral & E4 & F$-$D & dx lenient & 14/16 (.62--.98) & 16/16 (.79--1.00) & 0/2 & $-$0.12 & 0.500 & 1.000\\
\midrule
\multicolumn{10}{@{}l}{\emph{Secondary family (Holm over 4): SF$-$F on medication identity-set strict completion}}\\
Qwen & E0 & SF$-$F & med strict & 13/16 (.54--.96) & 10/16 (.35--.85) & 3/0 & +0.19 & 0.250 & 0.500\\
Qwen & E4 & SF$-$F & med strict & 14/16 (.62--.98) & 9/16 (.30--.80) & 5/0 & +0.31 & 0.0625 & 0.250\\
Mistral & E0 & SF$-$F & med strict & 15/16 (.70--1.00) & 10/16 (.35--.85) & 5/0 & +0.31 & 0.0625 & 0.250\\
Mistral & E4 & SF$-$F & med strict & 12/16 (.48--.93) & 11/16 (.41--.89) & 3/2 & +0.06 & 1.000 & 1.000\\
\bottomrule\end{tabular}
\par\smallskip\parbox{\linewidth}{\footnotesize succ.\ = successes over patient pairs with both cells scored in valid units; disc.\ A/B = discordant pairs A-only / B-only; diff = paired difference; $p$ = exact two-sided McNemar; Holm within the named family; exploratory at this sample; non-significance is not equivalence.}

\end{table}
\begin{table}[!tbp]\centering
\caption{Batch A under amendment 2 strict: SF vs F per model and condition (patient-paired exact McNemar; Holm-4; disc.\ = discordant pairs); amendment-1, contract and fidelity counts for reference, where SF/F gives the SF count and the F count of passing patients (two counts of 16, not a fraction).}\label{tab:batchAamend2}
\scriptsize\setlength{\tabcolsep}{1.5pt}\begin{tabular}{@{}llrrrrrrrrrr@{}}\toprule
model & cond. & SF (amend.\,2) & F (amend.\,2) & disc.\ SF-only/F-only & diff. & exact $p$ & Holm-4 $p$ & SF/F (amend.\,1) & SF/F (contract) & SF/F (fidelity)\\\midrule
Qwen & E0 & 13/16 & 13/16 & 1/1 & +0.000 & 1.000 & 1.000 & 13/13 & 13/13 & 11/11\\
Qwen & E4 & 11/16 & 13/16 & 0/2 & $-$0.125 & 0.500 & 1.000 & 11/13 & 11/13 & 10/11\\
Mistral & E0 & 4/16 & 10/16 & 0/6 & $-$0.375 & 0.031 & 0.125 & 4/10 & 3/10 & 6/8\\
Mistral & E4 & 4/16 & 10/16 & 1/7 & $-$0.375 & 0.070 & 0.211 & 4/10 & 3/10 & 7/8\\
\bottomrule\end{tabular}

\end{table}
\paragraph{Batch B.} The use calls were re-issued at temperature 0.2 with two seeds: 17/256 (Qwen) and 36/256 (Mistral) raw outputs changed between seeds, 1--2 of 128 fidelity-strict outcomes per model flipped against seed 0, and no primary cell is significant under fidelity (Holm $p=1.0$; Table~\ref{tab:batchB}) or amendment 2 (smallest Holm-8 $p=\vRuBatchBMinHolmStrict$; Table~\ref{tab:batchBamend2}).
\begin{table}[!tbp]\centering
\caption{Batch B: SF$-$F on fidelity-strict record fidelity per model, condition and seed (one Holm family of 8; disc.\ = discordant pairs). Medication strict completion, SF$-$F (Holm-8): seed 1 Qwen E0 13 vs 10 (Holm 1.000); seed 1 Qwen E4 13 vs 10 (Holm 1.000); seed 1 Mistral E0 14 vs 10 (Holm 1.000); seed 1 Mistral E4 11 vs 12 (Holm 1.000); seed 2 Qwen E0 13 vs 10 (Holm 1.000); seed 2 Qwen E4 14 vs 9 (Holm 0.500); seed 2 Mistral E0 14 vs 11 (Holm 1.000); seed 2 Mistral E4 12 vs 11 (Holm 1.000). Per-cell flips against seed 0 (problem list / medication): Qwen seed 1 1/128 / 6/128; Qwen seed 2 1/128 / 3/128; Mistral seed 1 2/128 / 5/128; Mistral seed 2 1/128 / 4/128.}\label{tab:batchB}
\footnotesize\begin{tabular}{@{}lllrrrrrr@{}}\toprule
seed & model & cond. & SF & F & disc.\ SF-only/F-only & diff. & exact $p$ & Holm-8 $p$\\\midrule
1 & Qwen & E0 & 11/16 & 11/16 & 1/1 & +0.000 & 1.000 & 1.000\\
1 & Qwen & E4 & 10/16 & 11/16 & 1/2 & $-$0.0625 & 1.000 & 1.000\\
1 & Mistral & E0 & 7/16 & 8/16 & 1/2 & $-$0.0625 & 1.000 & 1.000\\
1 & Mistral & E4 & 7/16 & 8/16 & 1/2 & $-$0.0625 & 1.000 & 1.000\\
2 & Qwen & E0 & 11/16 & 11/16 & 1/1 & +0.000 & 1.000 & 1.000\\
2 & Qwen & E4 & 10/16 & 11/16 & 1/2 & $-$0.0625 & 1.000 & 1.000\\
2 & Mistral & E0 & 7/16 & 8/16 & 1/2 & $-$0.0625 & 1.000 & 1.000\\
2 & Mistral & E4 & 7/16 & 8/16 & 1/2 & $-$0.0625 & 1.000 & 1.000\\
\bottomrule\end{tabular}

\end{table}
\begin{table}[!tbp]\centering
\caption{Batch B under amendment 2 strict per seed (Holm-8; disc.\ = discordant pairs); amendment-1, contract and fidelity counts for reference, where SF/F gives the SF count and the F count of passing patients (two counts of 16, not a fraction).}\label{tab:batchBamend2}
\scriptsize\setlength{\tabcolsep}{1pt}\resizebox{\textwidth}{!}{\begin{tabular}{@{}lllrrrrrrrrrr@{}}\toprule
seed & model & cond. & SF (amend.\,2) & F (amend.\,2) & disc.\ SF-only/F-only & diff. & exact $p$ & Holm-8 $p$ & SF/F (amend.\,1) & SF/F (contract) & SF/F (fidelity)\\\midrule
1 & Qwen & E0 & 13/16 & 13/16 & 1/1 & +0.000 & 1.000 & 1.000 & 13/13 & 13/13 & 11/11\\
1 & Qwen & E4 & 11/16 & 13/16 & 0/2 & $-$0.125 & 0.500 & 1.000 & 11/13 & 11/13 & 10/11\\
1 & Mistral & E0 & 4/16 & 10/16 & 1/7 & $-$0.375 & 0.070 & 0.562 & 4/10 & 3/10 & 7/8\\
1 & Mistral & E4 & 4/16 & 10/16 & 1/7 & $-$0.375 & 0.070 & 0.562 & 4/10 & 3/10 & 7/8\\
2 & Qwen & E0 & 13/16 & 13/16 & 1/1 & +0.000 & 1.000 & 1.000 & 13/13 & 13/13 & 11/11\\
2 & Qwen & E4 & 11/16 & 13/16 & 0/2 & $-$0.125 & 0.500 & 1.000 & 11/13 & 11/13 & 10/11\\
2 & Mistral & E0 & 5/16 & 10/16 & 1/6 & $-$0.312 & 0.125 & 0.625 & 5/10 & 4/10 & 7/8\\
2 & Mistral & E4 & 4/16 & 10/16 & 1/7 & $-$0.375 & 0.070 & 0.562 & 4/10 & 3/10 & 7/8\\
\bottomrule\end{tabular}
}
\end{table}

\FloatBarrier
\section{Replacement batch: eligibility and per-configuration counts}\label{app:batchC}\suppressfloats[t]
\begin{table}[!tbp]\centering
\caption{Replacement batch: 13 development patients per model, one decoding; configuration = pipeline, no-prop.\ = no-propagation; exact Clopper--Pearson 95\,\% intervals on success; patient-paired exact McNemar with Holm correction as pre-declared; both present = old and new value listed together. Development evidence; counterfactual field changes; non-significance is not equivalence.}\label{tab:batchC}
\footnotesize\setlength{\tabcolsep}{5pt}\begin{tabular}{@{}llrrrrr@{}}\toprule
model & configuration & success & new adopted & old retained & both present & others retained\\\midrule
Qwen & no-prop. & 0/13 (0.00--0.25) & 0/13 & 8/13 & 0/13 & 12/13\\
Qwen & quarantine & 0/13 (0.00--0.25) & 0/13 & 0/13 & 0/13 & 10/13\\
Qwen & re-read & 9/13 (0.39--0.91) & 13/13 & 0/13 & 0/13 & 9/13\\
Qwen & rebuild & 7/13 (0.25--0.81) & 8/13 & 0/13 & 0/13 & 12/13\\
Qwen & local repair & 8/13 (0.32--0.86) & 9/13 & 0/13 & 0/13 & 12/13\\
Mistral & no-prop. & 0/13 (0.00--0.25) & 0/13 & 8/13 & 0/13 & 11/13\\
Mistral & quarantine & 0/13 (0.00--0.25) & 0/13 & 0/13 & 0/13 & 10/13\\
Mistral & re-read & 10/13 (0.46--0.95) & 13/13 & 0/13 & 0/13 & 10/13\\
Mistral & rebuild & 8/13 (0.32--0.86) & 9/13 & 0/13 & 0/13 & 12/13\\
Mistral & local repair & 7/13 (0.25--0.81) & 8/13 & 0/13 & 0/13 & 12/13\\
\midrule
\multicolumn{7}{@{}p{0.97\linewidth}@{}}{\footnotesize primary (Holm-2), local repair vs re-read on success: Qwen 8 vs 9 (disc.\ 2/3, $p=1.000$, Holm 1.000); Mistral 7 vs 10 (disc.\ 2/5, $p=0.453$, Holm 0.906)}\\
\multicolumn{7}{@{}p{0.97\linewidth}@{}}{\footnotesize tertiary (Holm-8), old retained, local repair vs each: Qwen vs re-read 0 vs 0 (Holm 1.000); Qwen vs no-prop. 0 vs 8 (Holm 0.0625); Qwen vs quarantine 0 vs 0 (Holm 1.000); Qwen vs rebuild 0 vs 0 (Holm 1.000); Mistral vs re-read 0 vs 0 (Holm 1.000); Mistral vs no-prop. 0 vs 8 (Holm 0.0625); Mistral vs quarantine 0 vs 0 (Holm 1.000); Mistral vs rebuild 0 vs 0 (Holm 1.000)}\\
\bottomrule\end{tabular}

\end{table}
Eligibility was decided from the input record by the frozen field-level scorer before any model output: a medication line with exactly one explicit strength field and no unsupported numeric expression, whose drug identity is unique and unambiguous among the allowed lines; the new value replaces the strength number by $2\times$ (else $0.5\times$; all 13 pairs are $2\times$). Of 27 development patients considered, 13 carried an eligible line (16 requested; not padded); the 13 targets cover seven distinct line texts and six drug identities. Outcome distribution: no-propagation 8 old-retained and 5 missing per model; quarantine 13 missing; re-read 13 adopted; rebuild 8/9 adopted and 5/4 missing; local repair 9/8 adopted and 4/5 missing (Qwen/Mistral). No unit listed both versions of the target; no unit's target field was labelled an unsupported addition. Mean calls and tokens per trajectory: re-read 3 and 1.9k--2.2k; no-propagation and quarantine 12 and 6.3k--8.4k; local repair 15 and 7.8k--9.6k; rebuild 21 and 10.1k--12.5k.

\paragraph{Two readings that need care.} First, target-field labels say nothing about other fields; one Mistral patient's enoxaparin line was spelled ``ENOKAPARIN'' in all five pipelines, which the lexical rule counts as a missing plus an unsupported identity. Second, in all 18 rebuild/local-repair units that did not adopt the new value (9 per model), the new text was in the facts shown, and most of these units had also left that line out before the revision (Qwen 9 of 9, Mistral 7 of 9)---an association on five patients per model, not a causal attribution.

\paragraph{Secondary family (Holm-6).} Local repair against no-propagation, quarantine and rebuild on \emph{success} for both models, in one family of six two-sided exact McNemar tests (the first two are controls that adopted in no unit). Qwen: local repair vs no-propagation 8 vs 0 (Holm-6 $p=0.0469$); Qwen: local repair vs quarantine 8 vs 0 (Holm-6 $p=0.0469$); Qwen: local repair vs rebuild 8 vs 7 (discordant 1/0, Holm-6 $p=1.0$); Mistral: local repair vs no-propagation 7 vs 0 (Holm-6 $p=0.0625$); Mistral: local repair vs quarantine 7 vs 0 (Holm-6 $p=0.0625$); Mistral: local repair vs rebuild 7 vs 8 (discordant 0/1, Holm-6 $p=1.0$) (the family behind the \S\ref{sec:replace} values).

\FloatBarrier
\section{Confirmatory replacement run (batch C2): pre-specification and results}\label{app:batchCtwo}
\begin{figure}[!tbp]\centering
\includegraphics[width=\textwidth]{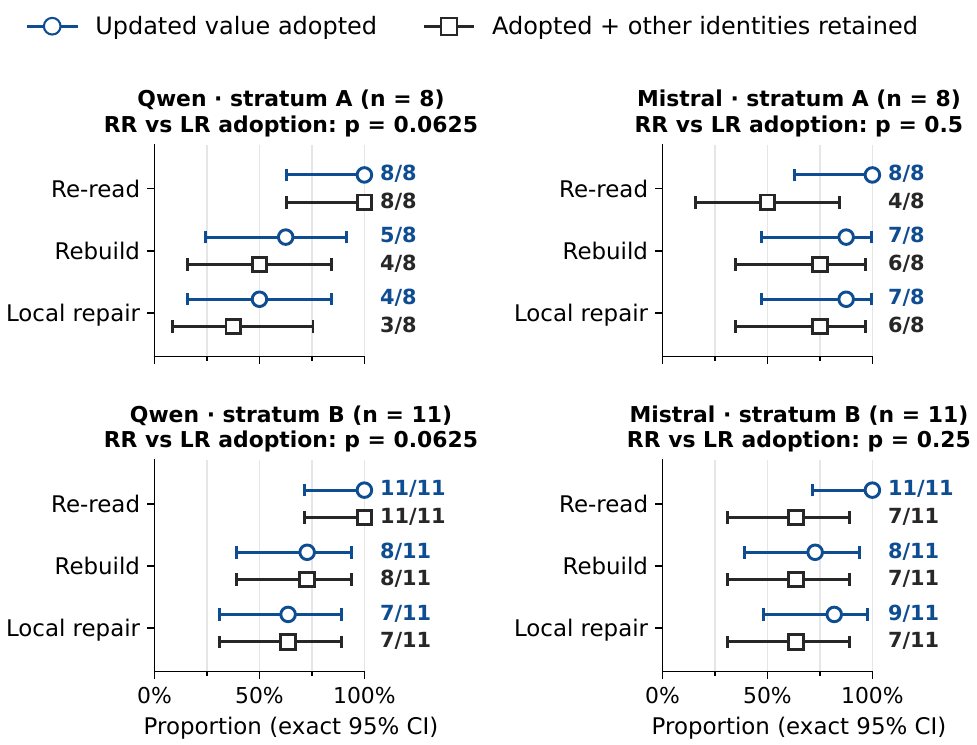}
\caption{C2 adoption and adoption with retention of all other valid medication identities, with exact 95\,\% Clopper--Pearson intervals. A: 8 previously unused protocol patients; B: 11 reused patients. Counts, models and strata are never pooled. Intervals describe individual proportions; displayed $p$ values use the paired one-sided adoption test, and no primary null hypothesis was rejected at $\alpha=0.05$. RR = re-reading, LR = local repair. No-propagation and quarantine have zero adoption; full counts and the stricter success definition appear in Table~\ref{tab:batchCtwo}.}\label{fig:c2}
\end{figure}

\suppressfloats[t]
\paragraph{Pre-specification.} The plan (hashed before any scenario was generated; amendment hashed before any model output) fixed: the same five batch-C pipelines and both models with identical snapshots and decoding (temperature 0, seed 0); a frozen build (the batch-C build plus partition and stage selection options), hash-listed before submission; the frozen field-level scorer; the patient as unit; primary hypothesis (labelled Q1 in the frozen plan), one-sided exact McNemar at $\alpha=0.05$: re-reading adopts the new value more often than local repair; secondary family (Holm-3 per model and stratum): re-reading versus rebuild on adoption (one-sided), and local repair and rebuild versus re-reading on success (two-sided); exact Clopper--Pearson intervals; no pooling across strata or models; the result is reported whatever its direction. \textbf{Plan amendment (before any output).} The plan intended the never-used sealed patients; the eligibility pass showed that all 20 generic-eligible sealed patients had been consumed by the held-out run, so the run was split into \emph{stratum A}: all 8 E2-eligible patients of the calibration partition (54 people; never used by any model run of this project, checked against every run directory; record-level exposure by the earlier project split of \S\ref{sec:setup} is not excluded), and \emph{stratum B}: the 11 E2-eligible held-out patients (second use of those records; disclosed as such). Eligibility was decided from the input record by the frozen rule of App.~\ref{app:batchC} (all pairs $2\times$). With $n=8$ the one-sided test at $\alpha=0.05$ rejects only when five or more discordant pairs all favour re-reading (5:0 to 8:0), or when all eight pairs are discordant and split 7:1; stratum A therefore has power only against a large effect and its interval, not its $p$-value, is the informative summary.
\paragraph{Execution.} One job per model (two GPU-hours reserved; one 80--96\,GB data-centre GPU each), stratum A then B on one served model, 40 and 55 trajectories per model; run status, code hash and scenario hash are recorded in the run directory and checked by the scoring script against the frozen list before scoring.
\paragraph{Results.} Table~\ref{tab:batchCtwo} and Fig.~\ref{fig:c2} give the counts; the tests follow.
\begin{table}[!tbp]\centering
\caption{Confirmatory replacement run: per-stratum, per-model counts (stratum A: 8 unused calibration patients; stratum B: 11 held-out patients, second use). No-prop.\ = no-propagation. Adopted = new strength value listed; old kept = superseded value listed; others ret.\ = every other valid medication identity retained; success = adopted $\wedge$ others retained; success$_{\text{strict}}$ additionally excludes unsupported identities. Strata are never pooled.}\label{tab:batchCtwo}
\footnotesize\begin{tabular}{@{}lllrrrrr@{}}\toprule
stratum & model & pipeline & adopted & old kept & others ret. & success & success$_{\text{strict}}$\\\midrule
A & Qwen & no-prop. & 0/8 & 5/8 & 7/8 & 0/8 & 0/8\\
A & Qwen & quarantine & 0/8 & 0/8 & 7/8 & 0/8 & 0/8\\
A & Qwen & re-read & 8/8 & 0/8 & 8/8 & 8/8 & 8/8\\
A & Qwen & rebuild & 5/8 & 0/8 & 7/8 & 4/8 & 4/8\\
A & Qwen & local repair & 4/8 & 0/8 & 6/8 & 3/8 & 3/8\\
\midrule
B & Qwen & no-prop. & 0/11 & 8/11 & 11/11 & 0/11 & 0/11\\
B & Qwen & quarantine & 0/11 & 0/11 & 11/11 & 0/11 & 0/11\\
B & Qwen & re-read & 11/11 & 0/11 & 11/11 & 11/11 & 11/11\\
B & Qwen & rebuild & 8/11 & 0/11 & 11/11 & 8/11 & 8/11\\
B & Qwen & local repair & 7/11 & 0/11 & 11/11 & 7/11 & 7/11\\
\midrule
A & Mistral & no-prop. & 0/8 & 6/8 & 7/8 & 0/8 & 0/8\\
A & Mistral & quarantine & 0/8 & 0/8 & 6/8 & 0/8 & 0/8\\
A & Mistral & re-read & 8/8 & 0/8 & 4/8 & 4/8 & 4/8\\
A & Mistral & rebuild & 7/8 & 0/8 & 7/8 & 6/8 & 6/8\\
A & Mistral & local repair & 7/8 & 0/8 & 7/8 & 6/8 & 6/8\\
\midrule
B & Mistral & no-prop. & 0/11 & 7/11 & 9/11 & 0/11 & 0/11\\
B & Mistral & quarantine & 0/11 & 0/11 & 9/11 & 0/11 & 0/11\\
B & Mistral & re-read & 11/11 & 0/11 & 7/11 & 7/11 & 7/11\\
B & Mistral & rebuild & 8/11 & 0/11 & 9/11 & 7/11 & 7/11\\
B & Mistral & local repair & 9/11 & 0/11 & 9/11 & 7/11 & 7/11\\
\bottomrule
\end{tabular}

\end{table}
\textbf{Stratum A, Qwen} (40/40 trajectories, 0 failed). Adoption: re-reading 8/8 (95\,\% CI [0.63, 1.00]), local repair 4/8 ([0.16, 0.84]), rebuild 5/8; primary one-sided test: discordant pairs 4 (re-read only) vs 0 (local only), $p=0.0625$, not rejected at $\alpha=0.05$. Secondary: re-read $>$ rebuild on adopted (one-sided): 8 vs 5, discordant 3/0, raw $p=0.125$, Holm-3 $p=0.250$; local repair vs re-read on success (two-sided): 3 vs 8, discordant 0/5, raw $p=0.0625$, Holm-3 $p=0.188$; rebuild vs re-read on success (two-sided): 4 vs 8, discordant 0/4, raw $p=0.125$, Holm-3 $p=0.250$. \textbf{Stratum A, Mistral} (40/40 trajectories, 0 failed). Adoption: re-reading 8/8 (95\,\% CI [0.63, 1.00]), local repair 7/8 ([0.47, 1.00]), rebuild 7/8; primary one-sided test: discordant pairs 1 (re-read only) vs 0 (local only), $p=0.500$, not rejected at $\alpha=0.05$. Secondary: re-read $>$ rebuild on adopted (one-sided): 8 vs 7, discordant 1/0, raw $p=0.500$, Holm-3 $p=1.000$; local repair vs re-read on success (two-sided): 6 vs 4, discordant 3/1, raw $p=0.625$, Holm-3 $p=1.000$; rebuild vs re-read on success (two-sided): 6 vs 4, discordant 3/1, raw $p=0.625$, Holm-3 $p=1.000$. \textbf{Stratum B, Qwen} (55/55 trajectories, 0 failed). Adoption: re-reading 11/11 (95\,\% CI [0.72, 1.00]), local repair 7/11 ([0.31, 0.89]), rebuild 8/11; primary one-sided test: discordant pairs 4 (re-read only) vs 0 (local only), $p=0.0625$, not rejected at $\alpha=0.05$. Secondary: re-read $>$ rebuild on adopted (one-sided): 11 vs 8, discordant 3/0, raw $p=0.125$, Holm-3 $p=0.375$; local repair vs re-read on success (two-sided): 7 vs 11, discordant 0/4, raw $p=0.125$, Holm-3 $p=0.375$; rebuild vs re-read on success (two-sided): 8 vs 11, discordant 0/3, raw $p=0.250$, Holm-3 $p=0.375$. \textbf{Stratum B, Mistral} (55/55 trajectories, 0 failed). Adoption: re-reading 11/11 (95\,\% CI [0.72, 1.00]), local repair 9/11 ([0.48, 0.98]), rebuild 8/11; primary one-sided test: discordant pairs 2 (re-read only) vs 0 (local only), $p=0.250$, not rejected at $\alpha=0.05$. Secondary: re-read $>$ rebuild on adopted (one-sided): 11 vs 8, discordant 3/0, raw $p=0.125$, Holm-3 $p=0.375$; local repair vs re-read on success (two-sided): 7 vs 7, discordant 3/3, raw $p=1.000$, Holm-3 $p=1.000$; rebuild vs re-read on success (two-sided): 7 vs 7, discordant 2/2, raw $p=1.000$, Holm-3 $p=1.000$. Every count, interval and patient list is recorded with the results.

\paragraph{Post-run protocol erratum (2026-09-18).} The frozen C2 plan describes two use calls and says that problem-list use is excluded. The frozen runner and all 190 recorded trajectories instead execute three calls: medication before revision, medication after revision, and a later problem-list use. Adoption and success are scored on the second call, so their definitions and reported counts are unchanged. Full-trajectory cost includes the additional call. This is a protocol--execution discrepancy, not an amendment made before the run; the original plan and hashes are retained.

\paragraph{Post-hoc pooled sign test.} Post hoc; contrary to the pre-specified no-pooling rule; reported for completeness only. Pooling the two strata within a model (patients are never pooled across models, which share them) gives 8 of 8 discordant pairs favouring re-reading for Qwen (one-sided sign test $p=0.0039$) and 3 of 3 for Mistral ($p=0.125$). The 11 pairs are 11 patient--model units, not 11 independent patients, and this is not a confirmatory result: the four pre-specified primary cells above are the confirmatory analysis and none rejected.

\FloatBarrier
\section{Adapted published repair selector}\label{app:adapted}\suppressfloats[t]
We implemented a \emph{paper-based adaptation} of the barrier-first cascade repair of \citet{zhao2026memorepair} as a configuration of our harness (\texttt{adapted selector}). As of September 2026 we found no author-released implementation (arXiv abstract and full text, and a public code search); this does not establish that none exists. The adaptation keeps the published structure: influence edges are our host-recorded parents; the cascade $C(F)$ is the invalidated root set $F$ (here the revised document) and its transitive successors; every member of $C(F)$ is withdrawn (barrier) before any operator runs; a deterministic mode map assigns \emph{recompute} to the claims of a corrected document, \emph{remove} to the claims of a revoked one and \emph{regen} to summaries with retained support; executability is checked; the selection maximises $\sum_i w_i x_i - \lambda \sum_i c_i x_i$ under predecessor closure, where $i$ ranges over the non-removed descendants, $x_i\in\{0,1\}$ marks a selected candidate, $w_i$ and $c_i$ are its value and cost (our proxies are given below) and $\lambda\ge 0$ is the trade-off weight (solved exactly by enumeration on our small cascades, where the paper uses a min-cut); unselected members stay withdrawn; a recomputed or regenerated node is republished only after a schema check and (our addition) a truth-free provenance check (its inputs are still active at the shown versions); leak, republication rate and normalised operator cost are recorded per revision.
Deviations, stated rather than emulated: semantic (retrieval) edges do not exist here and are not needed by the cascade; the \emph{migrate} event and the \emph{param} mode have no counterpart; the paper assigns modes by artefact class (\emph{recompute} for replayable deterministic artefacts), and we treat temperature-0 claim extraction as replayable; claims of a revoked document are assigned \emph{remove}, whereas the paper treats deterministic descendants of a deleted source as structurally non-executable candidates, which enter its republication-rate denominator; our operator costs are prompt characters over a predicted full-rebuild baseline, a prompt-length proxy, not tokens or the paper's Repair-all cost, so $\lambda=0.3$ matches the paper's value only numerically; the published paper defines the value $w_i$ and cost $c_i$ only as fixed value--cost terms, so ours are observable proxies (pre-event dependents $+1$; predicted prompt characters); the paper's task-regression and replay-consistency validators have no counterpart, and no experiment gold is read at repair time.
Deterministic tests (scripted clients, development scenarios): independent support survives an unrelated revocation (19 untouched nodes keep version and digest); the barrier equals the transitive reach; on the replacement graph the selected set is $\{c(\mathrm{D4}), s(\mathrm{D4}), G\}$ (node names as in Fig.~\ref{fig:pipeline}) for $\lambda\in\{0,0.3,1\}$, $\{c(\mathrm{D4}), s(\mathrm{D4})\}$ at $\lambda=10$, $\{c(\mathrm{D4})\}$ at 30 and $\emptyset$ at 100; a no-revision scenario triggers no operator. On these chain-shaped cascades the $\lambda=0.3$ of the paper's main configuration selects the same operator set as local repair (3 calls), so the two configurations differ only when validation fails (a malformed regenerated summary stays withdrawn under the adapted selector and is republished as raw text by local repair) or when $\lambda$ exceeds the per-node thresholds. Whether this changes real-model outcomes was tested in a pre-specified development batch whose results are reported in App.~\ref{app:batchE}.

\FloatBarrier
\section{Batches and cost}\label{app:repro}\suppressfloats[t]
\begin{table}[!tbp]\centering
\caption{Model-run batches that produced reported results, and the four failed 32B launches, with trajectories (traj.) and GPU-seconds (GPU-s). dev = development; calib.\ = calibration; local = local repair; no-prop.\ = no-propagation; $T$ = temperature; batch-A/B unit = one snapshot per patient, condition and model; L1--L3 = EXP-L1--L3. Role: dev.\ = development evidence; p.s.\ = a separate analysis plan hashed in advance, as stated in the relevant appendix; confirm.\ = pre-specified confirmatory test; expl.\ = exploratory 32B supplement, run after the 7B results were known under a hashed plan (App.~\ref{app:largemodel}). Pilots, smoke tests, cancelled jobs and the serving checks are not listed; see the reconciliation below. Patients are distinct people; the 8 development targets are shared by the screening and E1b/E4 batches; the 16 batch-A and 13 batch-C patients are development patients (7 in common); batch E uses the 13 batch-C patients and 4 of the 8 screening targets; 20 sealed patients were used in the original held-out run, 11 of them again for C2, and all 20 again by the 32B supplements (batch ledger in App.~\ref{app:largemodel}).}\label{tab:batchesfull}
\scriptsize\setlength{\tabcolsep}{3pt}\begin{tabular}{@{}>{\raggedright\arraybackslash}p{2.6cm}>{\raggedright\arraybackslash}p{2.55cm}l>{\raggedright\arraybackslash}p{3.4cm}rrl@{}}\toprule
batch & patients & events & pipelines / branches & traj. & GPU-s & role\\\midrule
screening (2 sizes) & 8 dev & E0, E1 & re-read, cache, rebuild, local & 256 & 4{,}206 & dev.\\
E1b/E4 controls (2 sizes) & same 8 dev & E1b, E4 & same four & 256 & 4{,}136 & dev.\\
held-out (pre-specified) & 20 sealed & E0, E1, E1b, E4 & same four & 640 & 5{,}050 & confirm.\\
batch A same-facts ablation & 16 dev & E0, E4 & F, SF, FS, D & 64 units & 878 & dev.\\
batch B decoding replay & same 16 dev & E0, E4 & same; $T{=}0.2$, seeds 1--2 & 128 units & 632 & dev.\\
batch C replacement & 13 dev & E2 & no-prop., quarantine, re-read, rebuild, local & 130 & 1{,}696 & dev.\\
batch E shared-ingest forks & 17 dev (21 units) & E2, E1 & re-read, rebuild, local, adapted $\lambda$=0.3, 10 & 210 & 1{,}818 & dev.\ (p.s.)\\
batch C2 confirmatory replacement & 8 calib.\ + 11 sealed (2nd use) & E2 & same five as batch C & 190 & 2{,}388 & confirm.\\
reuse sweep EXP-R (2 sizes; 20 uses) & same 8 dev & E1 & same four as screening & 128 & 5{,}014 & dev.\ (p.s.)\\
32B supplement L1 (2 sizes) & same 8 dev & E1 & same four; Qwen2.5-32B & 64 & 3{,}140 & expl.\\
32B replacement L3, dev13 & 13 dev (batch C) & E2 & same five as batch C; Qwen2.5-32B & 65 & 1{,}859 & expl.\\
32B replacement L3, calib8 & 8 calib.\ (2nd use) & E2 & same five; Qwen2.5-32B & 40 & 1{,}259 & expl.\\
32B replacement L3, sealed11 & 11 sealed (3rd use) & E2 & same five; Qwen2.5-32B & 55 & 1{,}811 & expl.\\
32B revocation L2, failed launches (4 jobs) & --- & --- & launcher error; no model output & 0 & 813 & ---\\
32B revocation L2 (E0, E1, E1b, E4; 4 jobs) & 6 calib.\ + 20 sealed (reused; App.~\ref{app:largemodel}) & E0, E1, E1b, E4 & same four as held-out; Qwen2.5-32B & 416 & 8{,}853 & expl.\\
length sweep EXP-X (4 levels; 20 uses) & same 8 dev (8, 3, 5, 6 per level) & E1 & same four + TF, RRTF & 264 & 11{,}629 & dev.\ (p.s.)\\
\bottomrule\end{tabular}

\end{table}
Each model-run batch used a frozen, hash-listed code build; builds advanced monotonically as options were added (partition and stage selection for the confirmatory replacement run; a repeated-use option and a model-registry entry for the reuse sweep and 32B supplement; length levels and two source-filtered use renderings for the length sweep), with no change to any scorer between them. The length-sweep plan records the Hugging Face snapshots of the two 7B tokenizers (Qwen2.5-7B-Instruct revision \texttt{a09a354}, Mistral-7B-Instruct-v0.3 revision \texttt{c170c70}) and states that every batch served the models at the same snapshot revisions. \citet{jiang2023mistral} describe the original Mistral~7B release and are cited for the model family; the v0.3 checkpoint is described in its Hugging Face model card \citep{mistral2024instructv03}. The problem-list scoring contract and its three amendments are post-hoc instruments, each hashed before the re-scoring it was applied to. Every batch had its scenarios, run configuration and the scoring contract of its pre-declared endpoints hashed before submission, and every job's completion state was recorded. Total GPU time of the 61 GPU jobs that consumed time is 73{,}137 GPU-seconds (20.32 GPU-hours; one 80--96\,GB data-centre GPU per job): 37{,}566 through C2 (pilots, smoke tests and cancelled jobs included), 337 for the 32B serving check, 5{,}014 for the reuse sweep, 3{,}140 for the 32B E1 supplement, 4{,}929 for the 32B replacement supplement, 813 for four 32B revocation launches that failed before producing output, 8{,}853 for the 32B revocation supplement, and 12{,}485 for the length sweep (11{,}629 batch, 382 serving check, 474 pilot). The batches listed in Table~\ref{tab:batchesfull} account for 55{,}182 of these GPU-seconds; the remaining 17{,}955 are the serving checks and the pilots, smoke tests and cancelled jobs that produced no reported result. Site charges are unverified. Code will be released upon acceptance; on release the archive will contain the code, frozen inputs, raw trajectories, scoring outputs, re-scoring diffs and audit scripts. Any anonymous evidence provided during review is a redacted derivative: text passed through a fixed identity map, run directories with user, account, host and home-path strings replaced by placeholders, and trajectory and result files unchanged.

\FloatBarrier
\section{Batch E: adapted selector from one shared ingest}\label{app:batchE}\suppressfloats[t]
Both runs completed (105 of 105 trajectories per model, 0 failed). Per model, 21 units (17 distinct development patients: 13 replacement units and 8 revocation units) were run through five pipelines; the four memory pipelines were forked from one shared ingested state per unit (byte-copied store; its hash is recorded in every row). Table~\ref{tab:batchE} gives the counts; the problem-list use of the same units is scored in Table~\ref{tab:batchEdx}.
\emph{Qwen.} Replacement adoption (of 13): re-read 13, rebuild 9, local repair 8, adapted selector ($\lambda$=0.3) 9, adapted selector ($\lambda$=10) 9. Replacement success, i.e.\ adopted and every other identity retained (of 13): re-read 9, rebuild 8, local repair 7, adapted selector ($\lambda$=0.3) 8, adapted selector ($\lambda$=10) 8. Old value retained: 0 in every pipeline. Revocation strict completion (of 8): re-read 6, rebuild 7, local repair 7, adapted selector ($\lambda$=0.3) 7, adapted selector ($\lambda$=10) 7; stale commits 0 of 8 in every pipeline; execution failures 0. Primary family (Holm-4): adapted selector ($\lambda$=0.3) vs local repair on replacement success: 8 vs 7 (discordant 1/0, Holm-4 $p$=1.000); adapted selector ($\lambda$=0.3) vs local repair on revocation strict completion: 7 vs 7 (discordant 0/0, Holm-4 $p$=1.000); adapted selector ($\lambda$=10) vs local repair on replacement success: 8 vs 7 (discordant 1/0, Holm-4 $p$=1.000); adapted selector ($\lambda$=10) vs local repair on revocation strict completion: 7 vs 7 (discordant 0/0, Holm-4 $p$=1.000). Secondary family (each memory pipeline vs re-read on the same endpoints; Holm-8): smallest adjusted $p$ = 1.000; no comparison is significant. Cost per trajectory, shared ingest counted for every memory pipeline: re-read 3.0 calls, 1{,}906 tokens; rebuild 20.2 calls, 10{,}021 tokens; local repair 14.2 calls, 7{,}679 tokens; adapted selector ($\lambda$=0.3) 14.2 calls, 7{,}679 tokens; adapted selector ($\lambda$=10) 13.2 calls, 6{,}768 tokens. Problem-list record fidelity under the original contract, not tabulated (of 21 units; Table~\ref{tab:batchEdx} gives fidelity and amendments 1 and 2): re-read strict 18, lenient 20; rebuild strict 14, lenient 19; local repair strict 14, lenient 19; adapted selector ($\lambda$=0.3) strict 14, lenient 19; adapted selector ($\lambda$=10) strict 13, lenient 17.
\emph{Mistral.} Replacement adoption (of 13): re-read 13, rebuild 10, local repair 10, adapted selector ($\lambda$=0.3) 10, adapted selector ($\lambda$=10) 8. Replacement success, i.e.\ adopted and every other identity retained (of 13): re-read 10, rebuild 9, local repair 8, adapted selector ($\lambda$=0.3) 8, adapted selector ($\lambda$=10) 6. Old value retained: 0 in every pipeline. Revocation strict completion (of 8): re-read 4, rebuild 5, local repair 5, adapted selector ($\lambda$=0.3) 5, adapted selector ($\lambda$=10) 4; stale commits 0 of 8 in every pipeline; execution failures 0. Primary family (Holm-4): adapted selector ($\lambda$=0.3) vs local repair on replacement success: 8 vs 8 (discordant 0/0, Holm-4 $p$=1.000); adapted selector ($\lambda$=0.3) vs local repair on revocation strict completion: 5 vs 5 (discordant 0/0, Holm-4 $p$=1.000); adapted selector ($\lambda$=10) vs local repair on replacement success: 6 vs 8 (discordant 0/2, Holm-4 $p$=1.000); adapted selector ($\lambda$=10) vs local repair on revocation strict completion: 4 vs 5 (discordant 0/1, Holm-4 $p$=1.000). Secondary family (each memory pipeline vs re-read on the same endpoints; Holm-8): smallest adjusted $p$ = 1.000; no comparison is significant. Cost per trajectory, shared ingest counted for every memory pipeline: re-read 3.0 calls, 2{,}121 tokens; rebuild 20.2 calls, 12{,}255 tokens; local repair 14.2 calls, 9{,}336 tokens; adapted selector ($\lambda$=0.3) 14.2 calls, 9{,}336 tokens; adapted selector ($\lambda$=10) 13.2 calls, 8{,}240 tokens. Problem-list record fidelity under the original contract, not tabulated (of 21 units; Table~\ref{tab:batchEdx} gives fidelity and amendments 1 and 2): re-read strict 19, lenient 21; rebuild strict 8, lenient 15; local repair strict 8, lenient 15; adapted selector ($\lambda$=0.3) strict 8, lenient 15; adapted selector ($\lambda$=10) strict 7, lenient 14.
\begin{table}[!tbp]\centering
\caption{Batch E (pre-specified development batch; 13 replacement and 8 revocation units per model from 17 distinct patients; memory pipelines forked from one shared ingest; one decoding). E2 success = new value adopted and other identities retained; calls and tokens are per trajectory with the shared ingest counted for every memory pipeline.}\label{tab:batchE}
\footnotesize\setlength{\tabcolsep}{3pt}\begin{tabular}{@{}llrrrrrrr@{}}\toprule
model & pipeline & E2 success & E2 adopted & E2 old kept & E1 strict & E1 stale & calls & tokens\\\midrule
Qwen & re-read & 9/13 & 13/13 & 0/13 & 6/8 & 0/8 & 3.0 & 1906\\
Qwen & rebuild & 8/13 & 9/13 & 0/13 & 7/8 & 0/8 & 20.2 & 10021\\
Qwen & local repair & 7/13 & 8/13 & 0/13 & 7/8 & 0/8 & 14.2 & 7679\\
Qwen & adapted selector ($\lambda=0.3$) & 8/13 & 9/13 & 0/13 & 7/8 & 0/8 & 14.2 & 7679\\
Qwen & adapted selector ($\lambda=10$) & 8/13 & 9/13 & 0/13 & 7/8 & 0/8 & 13.2 & 6768\\
\midrule
Mistral & re-read & 10/13 & 13/13 & 0/13 & 4/8 & 0/8 & 3.0 & 2121\\
Mistral & rebuild & 9/13 & 10/13 & 0/13 & 5/8 & 0/8 & 20.2 & 12255\\
Mistral & local repair & 8/13 & 10/13 & 0/13 & 5/8 & 0/8 & 14.2 & 9336\\
Mistral & adapted selector ($\lambda=0.3$) & 8/13 & 10/13 & 0/13 & 5/8 & 0/8 & 14.2 & 9336\\
Mistral & adapted selector ($\lambda=10$) & 6/13 & 8/13 & 0/13 & 4/8 & 0/8 & 13.2 & 8240\\
\bottomrule\end{tabular}

\end{table}
\begin{table}[!tbp]\centering
\caption{Batch E problem-list use under fidelity, amendment 1 and amendment 2 (strict and lenient): 21 condition--patient units per pipeline and model from 17 distinct patients (13 replacement + 8 revocation units); descriptive counts only, no test is computed on these pooled units.}\label{tab:batchEdx}
\footnotesize\begin{tabular}{@{}llrrrrrr@{}}\toprule
model & pipeline & \multicolumn{2}{c}{fidelity} & \multicolumn{2}{c}{amendment 1} & \multicolumn{2}{c}{amendment 2}\\
\cmidrule(lr){3-4}\cmidrule(lr){5-6}\cmidrule(lr){7-8}
 & & strict & lenient & strict & lenient & strict & lenient\\\midrule
Qwen & re-read & 15/21 & 20/21 & 18/21 & 20/21 & 18/21 & 20/21\\
Qwen & rebuild & 13/21 & 19/21 & 14/21 & 19/21 & 14/21 & 19/21\\
Qwen & local repair & 13/21 & 19/21 & 14/21 & 19/21 & 14/21 & 19/21\\
Qwen & adapted selector ($\lambda=0.3$) & 13/21 & 19/21 & 14/21 & 19/21 & 14/21 & 19/21\\
Qwen & adapted selector ($\lambda=10$) & 13/21 & 18/21 & 13/21 & 17/21 & 13/21 & 17/21\\
\midrule
Mistral & re-read & 16/21 & 21/21 & 19/21 & 21/21 & 19/21 & 21/21\\
Mistral & rebuild & 8/21 & 16/21 & 9/21 & 15/21 & 9/21 & 15/21\\
Mistral & local repair & 8/21 & 16/21 & 9/21 & 15/21 & 9/21 & 15/21\\
Mistral & adapted selector ($\lambda=0.3$) & 8/21 & 16/21 & 9/21 & 15/21 & 9/21 & 15/21\\
Mistral & adapted selector ($\lambda=10$) & 7/21 & 15/21 & 8/21 & 14/21 & 8/21 & 14/21\\
\bottomrule\end{tabular}

\end{table}%
The 8 revocation units come from 4 patients, so the revocation comparisons pair units, not patients, departing from the patient-as-unit rule of \S\ref{sec:task}; every adjusted $p$ in them is 1.0, and we read the batch-E tests as descriptive only.
\paragraph{Determinism check.} Pairing the $i$-th model call of the adapted selector ($\lambda=0.3$) with the $i$-th call of local repair in each unit (110 pairs per model): 7 (Qwen) and 4 (Mistral) pairs had byte-identical prompts and different responses; 8 and 2 later prompts differed as a consequence; the operator sets and call counts were identical in every unit.
\paragraph{Further results.} In all 31 memory-pipeline non-adoptions the new value was present verbatim in the use-stage prompt (Qwen 17, Mistral 14): omissions of visible current information, whose cause within the memory prompt this comparison does not isolate. No primary or secondary comparison was significant (Holm-4 and Holm-8 per model); stale commits were 0/8 in every revocation cell.
\paragraph{Selector behaviour.} The adapted selector at $\lambda=0.3$ selected exactly local repair's operators in every unit (same calls; logical token totals differ by 5 and 12), a structural consequence of our value and cost definitions (minimum $w_i/c_i$ 4.4 and 4.2); its outputs differed only where the server returned different text for byte-identical prompts at temperature 0 (7 and 4 of 110 paired calls). At $\lambda=10$ it left the global summary withdrawn, saved one call per revision, and adopted 9/13 and 8/13.

\FloatBarrier
\section{Reuse sweep (EXP-R)}\label{app:reuse}
\begin{figure}[!tbp]\centering\includegraphics[width=\textwidth]{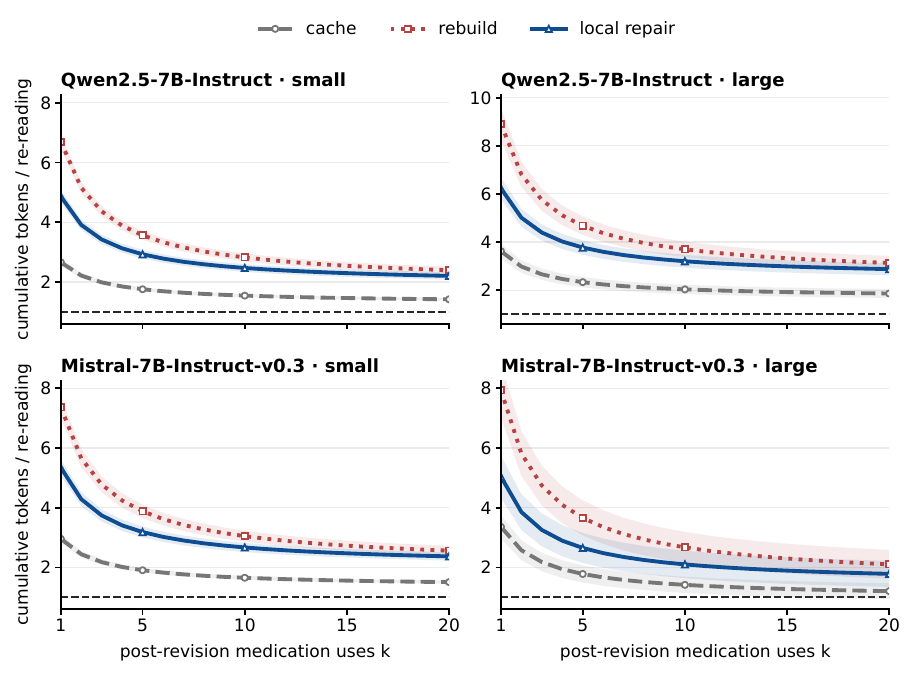}
\caption{Reuse sweep: cumulative tokens of each memory pipeline relative to re-reading, $C_{\mathrm{mem}}(k)/C_{\mathrm{reread}}(k)$, for $k=1$ to $20$ later uses (mean over eight paired development patients with 95\,\% $t$-intervals; one panel per model and size; dashed line at 1.0). No mean curve reaches 1.0.}\label{fig:reuse}\end{figure}

\suppressfloats[t]
\paragraph{Pre-specification and execution.} The sweep was run under a pre-specification plan (hashed before any model call under it) and a frozen build (the confirmatory-run build plus a repeated-use option and a model-registry entry, no scoring change). Units: the eight development screening patients of Table~\ref{tab:cost}, both sizes, E1, the four pipelines and both 7B models (128 trajectories, 0 execution failures). After the E1 revision and use$_1$ each trajectory issued 19 further medication uses (use$_2$ to use$_{20}$) with the identical use-prompt template; problem-list uses were not issued. Prompt+completion tokens per use were taken from the server usage record. One 80--96\,GB data-centre GPU per model, 2{,}412 and 2{,}602 GPU-seconds. Execution amendment (dated before the jobs started): the jobs were first submitted to the default queue, cancelled before starting when the scheduler gave no start estimate, and resubmitted unchanged to a short-job queue. No selective re-run occurred.
\paragraph{Analysis (measured, no extrapolation).} Per patient, $C(k)=\text{ingest}+\text{use}_0+\text{revise}+\sum_{j\le k}\text{use}_j$ for the memory pipeline and for re-reading on the same patient; $\Delta C(k)$ is the paired difference and Fig.~\ref{fig:reuse} reports the mean of the per-patient ratio $C_{\mathrm{mem}}(k)/C_{\mathrm{reread}}(k)$ (a mean of paired ratios, not a ratio of mean costs) with a 95\,\% $t$-interval over the eight patients. Cumulative $\Delta C(k)$ did not cross zero within $k\le 20$ in any of the 12 cells. The mean per-use difference $U_{\mathrm{mem}}-U_{\mathrm{reread}}$ over uses 1--20 was positive in 11 cells (+175 to +2{,}276 tokens) and negative in one (Mistral-7B, large, cache: $-12.6$ tokens per use against a fixed-cost difference $\Delta I+\Delta U_0+\Delta R=+8{,}213$; the algebraic break-even $8{,}213/12.6\approx 652$ uses is a quotient of two measured means, not an observed crossing). For the same cell the two later uses of the original screening run (Table~\ref{tab:cost}) had given $-57$ tokens per use; the 20-use sweep measured $-12.6$. The sweep is a separate run of the same scenarios, so its ingest and use$_0$ differ slightly from the screening run's: its $\Delta U_0$ for this cell is $-27$ and its fixed-cost difference 8{,}213, against $-73$ and 8{,}329 in Table~\ref{tab:cost}; \S\ref{sec:cost} quotes the sweep's values. The local-repair later-use ratio over uses 1--20 (mean memory use tokens over mean re-reading use tokens) is 2.51, 1.91, 1.42 and 2.03 for Qwen large, Qwen small, Mistral large and Mistral small. Table~\ref{tab:reuse} gives the cumulative ratios at $k=1$, 5, 10 and 20 and the per-use differences; full per-$k$ values will be released with the code.
\begin{table}[!tbp]\centering
\caption{Reuse sweep: cumulative cost relative to re-reading at $k=1,5,10,20$ (mean [95\,\% $t$-interval] over eight paired patients) and the mean per-use token difference $U_{\mathrm{mem}}-U_{\mathrm{reread}}$ over uses 1--20.}\label{tab:reuse}
\fontsize{8}{9.5}\selectfont\setlength{\tabcolsep}{2.5pt}\begin{tabular}{@{}lllrrrrr@{}}\toprule
model & size & pipeline & $k{=}1$ & $k{=}5$ & $k{=}10$ & $k{=}20$ & per-use $\Delta$ (tokens)\\\midrule
Qwen & large & cache & 3.62 [3.31, 3.92] & 2.33 [2.12, 2.54] & 2.03 [1.84, 2.23] & 1.86 [1.68, 2.04] & +1036.6\\
Qwen & large & local repair & 6.23 [5.78, 6.68] & 3.77 [3.47, 4.07] & 3.20 [2.93, 3.47] & 2.87 [2.62, 3.12] & +2275.5\\
Qwen & large & rebuild & 8.90 [8.23, 9.57] & 4.68 [4.30, 5.06] & 3.70 [3.37, 4.02] & 3.13 [2.84, 3.42] & +2265.4\\
Qwen & small & cache & 2.66 [2.54, 2.78] & 1.76 [1.69, 1.82] & 1.54 [1.49, 1.60] & 1.42 [1.37, 1.47] & +175.3\\
Qwen & small & local repair & 4.86 [4.66, 5.07] & 2.93 [2.78, 3.07] & 2.47 [2.34, 2.60] & 2.20 [2.08, 2.33] & +568.0\\
Qwen & small & rebuild & 6.69 [6.46, 6.93] & 3.56 [3.42, 3.71] & 2.82 [2.69, 2.95] & 2.39 [2.28, 2.51] & +573.2\\
Mistral & large & cache & 3.36 [2.97, 3.74] & 1.78 [1.49, 2.07] & 1.42 [1.14, 1.69] & 1.21 [0.94, 1.47] & $-$12.6\\
Mistral & large & local repair & 5.03 [4.33, 5.74] & 2.65 [2.15, 3.15] & 2.10 [1.64, 2.56] & 1.78 [1.35, 2.22] & +715.6\\
Mistral & large & rebuild & 7.93 [7.02, 8.85] & 3.66 [3.07, 4.24] & 2.67 [2.15, 3.19] & 2.10 [1.62, 2.59] & +780.8\\
Mistral & small & cache & 2.96 [2.82, 3.10] & 1.91 [1.82, 1.99] & 1.65 [1.59, 1.72] & 1.51 [1.45, 1.56] & +231.4\\
Mistral & small & local repair & 5.34 [5.06, 5.62] & 3.18 [3.00, 3.36] & 2.67 [2.52, 2.82] & 2.37 [2.23, 2.51] & +696.4\\
Mistral & small & rebuild & 7.36 [6.99, 7.73] & 3.88 [3.66, 4.10] & 3.05 [2.87, 3.23] & 2.56 [2.40, 2.72] & +686.6\\
\bottomrule
\end{tabular}

\end{table}

\FloatBarrier
\section{Length sweep (EXP-X): pre-specification, amendments and full results}\label{app:length}\suppressfloats[t]
\paragraph{Pre-specification.} A plan, a frozen build, a scenario manifest (22 scenario files) and the analysis program were hashed before the first model call of the batch. The build is the reuse-sweep build plus a length-level scenario builder and the TF and RRTF use renderings; the small-scenario builder is byte-identical, so level S is the scenario of the main batches. TF maintains memory exactly as local repair does and, at use time, shows only the claims whose parent document has an allowed source, with those documents' chunk summaries and without the global summary; RRTF is re-reading restricted to allowed-source documents. Units: the eight development screening patients of Table~\ref{tab:cost}, condition E1, seed 0, one greedy decoding for every stage, and 20 later medication uses with the identical template (no problem-list use). The vLLM context limit was raised to 32{,}768 tokens; a serving check sent the longest XL re-reading prompt (9{,}825 Qwen and 10{,}361 Mistral tokens) twice to each model and received identical, strictly parsed answers. A level is the token count (Qwen2.5 tokenizer) of the post-revision re-reading document body, headers and small-scenario documents included. Context documents are the patient's own records from the eICU demo tables \texttt{lab}, \texttt{vitalperiodic}, \texttt{vitalaperiodic} and \texttt{nursecharting}, one document per table and ICU day, with connector source \texttt{lab} or \texttt{ehr}; the medication task allows only pharmacy sources, so no context document can support a medication key. The intake--output table was excluded because its labels carry drug names; a scan of all 275 medication keys found none in the vital-sign tables and only analyte names (sodium, calcium, potassium, glucose) in the laboratory and nursing tables, the same collision the small scenario's laboratory document already contains. Context documents are appended after the small documents and the revision event, so D1--D4 and the revision are byte-identical to the small scenario. They are ordered chronologically across tables (fixed table order within one ICU day) and added whole while the running total stays within the target, until it reaches 90\,\% of the target; the rule reads token counts only. A patient enters a level if its re-reading body lies within $[0.9, 1.0]\times$ the target; at S these bodies are 251--373 Qwen2.5 tokens (Mistral 276--419). XL was drafted at 12{,}000 tokens and set to 10{,}000 by the plan's own fallback rule (the largest multiple of 1{,}000 reached by at least six of the eight patients). This gave 8, 3, 5 and 6 patients at S, M, L and XL; one patient is in band at all of M, L and XL, so each level is analysed on its own patients and M ($n=3$) is descriptive.

\paragraph{Amendments and execution.} Two revisions were made before the plan was hashed and before any model call: the drafted fill rule, which stopped at the first document that would cross the target and reached the band for none of the eight patients at M, became the skip-and-continue rule above, and documents were ordered chronologically across tables instead of table by table, which had left fewer patients in band. After hashing, a scheduler amendment fixed explicit queue options; after the pilot job was submitted but before its projection was recorded or any batch job was submitted, a logistics amendment raised the GPU budget cap from 20 to 60 GPU-hours and moved the collection deadline. A one-patient XL pilot on Qwen (not a result; never analysed) projected 3.85 GPU-hours for the batch (about 4.1 with the serving check and the pilot), within the original cap, so the raise was not needed. Neither amendment changed the design, the scenarios or the decoding. The batch ran as eight jobs (model $\times$ level): 264 trajectories, no execution failure, no duplicate and no resubmission, in 11{,}629 GPU-seconds (one 80--96\,GB data-centre GPU per job; 12{,}485 with the serving check and the pilot). A seventh, optional pipeline (stateless re-extraction at every use) was declined before the freeze.

\paragraph{Endpoints and a disclosed omission.} Tokens are prompt plus completion from the server usage record. Per patient, the per-use difference and the per-use ratio average uses 1--20; cells report the mean over patients with a 95\,\% $t$-interval. $k^*_{\mathrm{obs}}$ is the first $k\le 20$ at which the patient-mean cumulative difference is below zero; $k^*_{\mathrm{pred}}=(\Delta I+\Delta U_0+\Delta R)/(U_{\mathrm{reread}}-U_{\mathrm{mem}})$ uses the per-use difference of uses 1--3 only (0 if the pipeline is cheaper from the start). Quality endpoints use exact McNemar tests against re-reading with Holm correction over the pipelines within each model and level; no direction was pre-specified for them. The cached-input sensitivity bills input tokens that repeat the previous call's prefix in the same trajectory at a fraction $c$ of an uncached token (Table~\ref{tab:lengthcached}). The analysis program counted the repeated prefix with the Qwen2.5 tokenizer for both models (its re-reading cached counts are identical for the two models), so Mistral's cached discounts are understated and its cached ratios favour memory. We therefore exclude Mistral's cached ratios from every claim; an exact recount needs the recorded prompts and Mistral's own tokenizer and has not been done. For scale only: multiplying Mistral's cached counts by its re-reading token ratio to Qwen at XL (1.06) suggests that the error could move Mistral's local repair from below to above full re-reading at $c=0.1$; this is a rough approximation, not a correction, and nothing in the paper relies on it. Table~\ref{tab:lengthcached} reproduces the hashed output except that Mistral's invalid cached ratios are replaced by dashes; the no-cache column, taken from server usage, is unaffected. The hashed analysis program printed four of the five pre-specified checks under shifted labels (its P1--P4 are the plan's P1, P3, P4 and P5, under a heading that cites the wrong plan section) and did not compute P2. Neither file was edited; P2 was computed afterwards, with the plan's wording, from the per-cell values the program wrote (Table~\ref{tab:lengthpred}). Because each level has its own patients, the S-to-M step in P2 is not a paired comparison. The plan's other endpoints are in Table~\ref{tab:lengthendpoints}: cumulative ratios at $k=1$, 5, 10 and 20, strict completion with exact Clopper--Pearson intervals, and the committed endpoint, which is $n$ of $n$ in every cell.
\begin{table}[!tbp]\centering
\caption{Length sweep: the five pre-specified predictions and their outcomes (8, 3, 5 and 6 patients at S, M, L and XL). P1 and P3--P5 are the analysis program's checks; P2 was computed afterwards from its per-cell output. RR = re-reading; C = cache; RB = rebuild; LR = local repair; TF = task-filtered local repair; RRTF = re-reading restricted to allowed sources.}\label{tab:lengthpred}
\footnotesize\setlength{\tabcolsep}{3pt}
\begin{tabular}{@{}l>{\raggedright\arraybackslash}p{5.6cm}>{\raggedright\arraybackslash}p{5.9cm}@{}}\toprule
 & pre-specified prediction & outcome\\\midrule
P1 & At S, the per-use difference is positive for C, LR and RB in both models. & Holds: Qwen $+176$, $+568$, $+580$; Mistral $+230$, $+681$, $+689$ tokens per use.\\
P2 & For C, LR and RB, the per-use ratio $U_{\mathrm{mem}}/U_{\mathrm{reread}}$ decreases from S to XL in both models. & Holds for S versus XL in 6 of 6 cells (Qwen 1.28--1.93 to 0.50--0.65; Mistral 1.34--2.03 to 0.14--0.23); a monotone reading fails for Qwen, whose M cell is above S for all three. Computed after the run (see text).\\
P3 & At XL, the per-use difference is negative for C and LR in at least one model. & Holds in both models, and for RB; every 95\,\% interval excludes zero.\\
P4 & Where both exist, $|k^*_{\mathrm{pred}}-k^*_{\mathrm{obs}}|\le\max(2, 0.25\,k^*_{\mathrm{obs}})$. & Holds in 23 of 23 cells: 15 cells with a crossing after use 1 (largest error 0.9 uses) and 8 RRTF cells cheaper from the start ($k^*_{\mathrm{pred}}=0$, $k^*_{\mathrm{obs}}=1$).\\
P5 & RRTF has the lowest cumulative cost at $k=20$ at every level in both models. & Holds in 8 of 8 model--level cells; largely implied by the design, which added only non-allowed documents.\\\bottomrule
\end{tabular}
\end{table}
\paragraph{Plan-specified subset reports.} The plan also asks for level S on each longer level's patients and for the one patient in band at M, L and XL. The hashed program, run unchanged on those patient subsets, gave Table~\ref{tab:lengthsubsets}. Restricting S to a level's patients moved re-reading's per-use tokens by at most 35 and the per-use ratios by at most 0.07, so the S-to-XL changes do not come from which patients were in band. On the same patients, the ratios of Qwen's cache, rebuild and local repair rose from S to M (1.31, 1.86 and 1.86 to 1.45, 2.09 and 2.08) before falling, and on the single patient in band at M, L and XL (program output, not tabulated) they also rose from S to M before falling (local repair 1.81, 2.21, 1.52 and 0.68 at S, M, L and XL); every Mistral ratio fell at each step. Paired on the XL patients, all six model--pipeline ratios fell from S to XL (Qwen 1.30--1.93 to 0.50--0.65; Mistral 1.35--2.04 to 0.14--0.23).
\begin{table}[!tbp]\centering
\caption{Length sweep, plan-specified subset report: re-reading tokens per use and per-use ratio to re-reading at S on all eight patients and on each longer level's own patients, next to that level's value (hashed program run unchanged on patient subsets; descriptive). RR, TF and RRTF as in Table~\ref{tab:lengthpred}.}\label{tab:lengthsubsets}
\footnotesize\setlength{\tabcolsep}{3.5pt}\begin{tabular}{@{}llrrrrrrr@{}}\toprule
 & & S & \multicolumn{2}{c}{M patients ($n=3$)} & \multicolumn{2}{c}{L patients ($n=5$)} & \multicolumn{2}{c}{XL patients ($n=6$)}\\
\cmidrule(lr){4-5}\cmidrule(lr){6-7}\cmidrule(l){8-9}
model & pipeline & all 8 & S & M & S & L & S & XL\\\midrule
Qwen & RR tokens/use & 623 & 592 & 1{,}232 & 652 & 4{,}156 & 630 & 9{,}584\\
 & cache & 1.28 & 1.31 & 1.45 & 1.28 & 0.95 & 1.30 & 0.50\\
 & rebuild & 1.93 & 1.86 & 2.09 & 1.93 & 1.30 & 1.93 & 0.64\\
 & local repair & 1.91 & 1.86 & 2.08 & 1.90 & 1.29 & 1.93 & 0.65\\
 & TF & 1.00 & 0.96 & 0.46 & 0.98 & 0.15 & 1.00 & 0.07\\
 & RRTF & 0.71 & 0.68 & 0.33 & 0.71 & 0.11 & 0.71 & 0.05\\
\midrule
Mistral & RR tokens/use & 678 & 647 & 1{,}354 & 713 & 4{,}445 & 696 & 10{,}130\\
 & cache & 1.34 & 1.38 & 0.87 & 1.34 & 0.34 & 1.35 & 0.14\\
 & rebuild & 2.03 & 2.07 & 1.37 & 2.02 & 0.53 & 2.04 & 0.23\\
 & local repair & 2.01 & 2.04 & 1.37 & 2.02 & 0.52 & 2.02 & 0.23\\
 & TF & 1.05 & 0.99 & 0.47 & 1.03 & 0.16 & 1.04 & 0.07\\
 & RRTF & 0.70 & 0.67 & 0.32 & 0.69 & 0.11 & 0.69 & 0.05\\
\bottomrule\end{tabular}

\end{table}
\paragraph{Full results.} Table~\ref{tab:lengthfull} gives every cell as written by the hashed analysis program, Table~\ref{tab:lengthcached} the pre-specified cached-input pricing and Table~\ref{tab:lengthendpoints} the remaining pre-specified endpoints.
\paragraph{Extraction at the output cap (exploratory).} Not a pre-specified endpoint; the cap criterion is that of App.~\ref{app:rho}. Of the ingest and revise extraction calls of the four memory pipelines, none of the 836 on the original documents stopped at the 2{,}048-token output cap, and none at M, but 225 of the 520 calls on added documents at L and XL did (Qwen 110 of 260, Mistral 115 of 260; Table~\ref{tab:lengthextract}); the server's finish reason and a completion count equal to the cap give the same counts in every model--level--pipeline cell. These 520 calls averaged 1{,}492 completion tokens (cell means 1{,}393--1{,}748), against 205--382 for extractions of the original documents; in the held-out run re-reading's three answers totalled 294--355 completion tokens (Table~\ref{tab:heldoutphases}). All 44 extractions that failed to parse after the labelled repair were Mistral's, on added documents at L and XL; Mistral's XL cache, rebuild and local repair used 31--37\,\% of Qwen's tokens per use (Table~\ref{tab:lengthfull}), and 38--43\,\% at L, where only 9 extractions failed to parse, so parse failures alone do not explain the difference. A complete extraction would have produced more ingest output and more rendered claims, so the break-evens at L and XL probably favour memory; this net effect, including any change in summary length, was not measured. A truncated or unparsed extraction admits fewer claims, so part of the memory pipelines' per-use saving at L and XL reflects content that never entered memory. The added documents carry no medication order and support no expected answer, so the medication endpoints do not depend on them; a task that used them would.
\begin{landscape}
\begin{table}\centering
\caption{Length sweep, full results as written by the hashed analysis program (Qwen here, Mistral on the next page; rows unchanged, split only for layout). Tokens are means over patients (prompt+completion); @$k$ = after $k$ later uses; per use = mean of uses 1--20; per-use $\Delta$ (pipeline $-$ RR) and ratio@20 are means of per-patient paired values with 95\,\% $t$-intervals against RR; use$_0$ = initial use; pipeline abbreviations as in Table~\ref{tab:lengthpred}; $k^*$ as defined in the text (observed/predicted, the prediction rounded to the nearest use; ``none'': not reached within 20 uses, or no negative per-use difference); strict = strict completion after the revision. Each column is rounded separately, so ingest + use$_0$ + revise + 20 $\times$ per use can differ from total@20 by rounding. M rows have $n=3$ and are descriptive. No trajectory listed the revoked key or a non-allowed item (stale and unsupported-source listings 0 in every cell); no strict-completion comparison with RR was significant (Holm $p=1.0$ in every cell).}\label{tab:lengthfull}
\small\setlength{\tabcolsep}{3pt}\begin{tabular}{@{}lllrrrrrrrrrr@{}}\toprule
model & level & pipeline & $n$ & ingest & use$_0$ & revise & per use & total@20 & per-use $\Delta$ [95\% CI] & ratio@20 [95\% CI] & $k^*_{\mathrm{obs}}$/$k^*_{\mathrm{pred}}$ & strict\\\midrule
Qwen & S & RR & 8 & 0 & 684 & 0 & 623 & 13142 & --- & 1.00 & --- & 6/8\\
Qwen & S & C & 8 & 1821 & 848 & 0 & 798 & 18639 & +176 [+153, +198] & 1.42 [1.37, 1.47] & none/none & 4/8\\
Qwen & S & RB & 8 & 3362 & 1280 & 2910 & 1202 & 31604 & +580 [+492, +667] & 2.40 [2.31, 2.50] & none/none & 7/8\\
Qwen & S & LR & 8 & 3366 & 1285 & 508 & 1191 & 28978 & +568 [+473, +663] & 2.21 [2.08, 2.33] & none/none & 7/8\\
Qwen & S & TF & 8 & 3366 & 702 & 507 & 623 & 17040 & +0 [$-$61, +62] & 1.30 [1.20, 1.39] & none/none & 8/8\\
Qwen & S & RRTF & 8 & 0 & 505 & 0 & 446 & 9425 & $-$177 [$-$209, $-$145] & 0.72 [0.66, 0.77] & 1/0 & 8/8\\
Qwen & M & RR & 3 & 0 & 1289 & 0 & 1232 & 25922 & --- & 1.00 & --- & 3/3\\
Qwen & M & C & 3 & 4747 & 1821 & 0 & 1784 & 42248 & +552 [+56, +1049] & 1.63 [1.18, 2.09] & none/none & 3/3\\
Qwen & M & RB & 3 & 7646 & 2528 & 7300 & 2558 & 68641 & +1327 [+805, +1848] & 2.66 [1.95, 3.37] & none/none & 3/3\\
Qwen & M & LR & 3 & 7662 & 2536 & 876 & 2554 & 62160 & +1323 [+819, +1827] & 2.41 [1.78, 3.04] & none/none & 3/3\\
Qwen & M & TF & 3 & 7667 & 642 & 786 & 574 & 20582 & $-$657 [$-$807, $-$508] & 0.79 [0.69, 0.90] & 12/12 & 3/3\\
Qwen & M & RRTF & 3 & 0 & 463 & 0 & 406 & 8590 & $-$825 [$-$868, $-$783] & 0.33 [0.22, 0.44] & 1/0 & 3/3\\
Qwen & L & RR & 5 & 0 & 4204 & 0 & 4156 & 87333 & --- & 1.00 & --- & 5/5\\
Qwen & L & C & 5 & 11908 & 3977 & 0 & 3933 & 94548 & $-$223 [$-$1011, +564] & 1.08 [0.87, 1.29] & none/51 & 4/5\\
Qwen & L & RB & 5 & 17081 & 5387 & 16736 & 5420 & 147600 & +1263 [+451, +2076] & 1.69 [1.45, 1.93] & none/none & 5/5\\
Qwen & L & LR & 5 & 17157 & 5451 & 1544 & 5371 & 131568 & +1214 [+385, +2043] & 1.51 [1.28, 1.73] & none/none & 5/5\\
Qwen & L & TF & 5 & 17039 & 721 & 1507 & 639 & 32043 & $-$3518 [$-$3748, $-$3288] & 0.37 [0.34, 0.40] & 5/4 & 5/5\\
Qwen & L & RRTF & 5 & 0 & 520 & 0 & 462 & 9756 & $-$3695 [$-$3873, $-$3517] & 0.11 [0.09, 0.14] & 1/0 & 5/5\\
Qwen & XL & RR & 6 & 0 & 9665 & 0 & 9584 & 201341 & --- & 1.00 & --- & 4/6\\
Qwen & XL & C & 6 & 19199 & 4795 & 0 & 4740 & 118804 & $-$4843 [$-$5998, $-$3689] & 0.59 [0.47, 0.72] & 3/3 & 5/6\\
Qwen & XL & RB & 6 & 24992 & 6283 & 24539 & 6170 & 179214 & $-$3414 [$-$4481, $-$2347] & 0.89 [0.76, 1.03] & 14/14 & 6/6\\
Qwen & XL & LR & 6 & 24976 & 6271 & 1543 & 6196 & 156710 & $-$3388 [$-$4493, $-$2283] & 0.78 [0.66, 0.90] & 7/7 & 6/6\\
Qwen & XL & TF & 6 & 25026 & 718 & 1509 & 635 & 39961 & $-$8948 [$-$9101, $-$8795] & 0.20 [0.18, 0.21] & 2/2 & 6/6\\
Qwen & XL & RRTF & 6 & 0 & 507 & 0 & 448 & 9477 & $-$9135 [$-$9297, $-$8973] & 0.05 [0.04, 0.05] & 1/0 & 6/6\\
\bottomrule
\end{tabular}

\end{table}
\begin{table}\ContinuedFloat\centering
\caption{Length sweep, full results (continued): Mistral. Columns and conventions as on the previous page.}
\small\setlength{\tabcolsep}{3pt}\begin{tabular}{@{}lllrrrrrrrrrr@{}}\toprule
model & level & pipeline & $n$ & ingest & use$_0$ & revise & per use & total@20 & per-use $\Delta$ [95\% CI] & ratio@20 [95\% CI] & $k^*_{\mathrm{obs}}$/$k^*_{\mathrm{pred}}$ & strict\\\midrule
Mistral & S & RR & 8 & 0 & 762 & 0 & 678 & 14320 & --- & 1.00 & --- & 4/8\\
Mistral & S & C & 8 & 2365 & 963 & 0 & 908 & 21484 & +230 [+209, +251] & 1.51 [1.45, 1.56] & none/none & 5/8\\
Mistral & S & RB & 8 & 4126 & 1501 & 3551 & 1367 & 36517 & +689 [+623, +755] & 2.56 [2.42, 2.70] & none/none & 4/8\\
Mistral & S & LR & 8 & 4122 & 1492 & 610 & 1359 & 33404 & +681 [+610, +753] & 2.34 [2.22, 2.47] & none/none & 5/8\\
Mistral & S & TF & 8 & 4132 & 809 & 622 & 709 & 19743 & +31 [$-$22, +85] & 1.38 [1.30, 1.46] & none/none & 5/8\\
Mistral & S & RRTF & 8 & 0 & 553 & 0 & 477 & 10091 & $-$201 [$-$237, $-$165] & 0.70 [0.65, 0.76] & 1/0 & 7/8\\
Mistral & M & RR & 3 & 0 & 1409 & 0 & 1354 & 28489 & --- & 1.00 & --- & 3/3\\
Mistral & M & C & 3 & 6210 & 1231 & 0 & 1169 & 30819 & $-$185 [$-$1019, +649] & 1.09 [0.40, 1.77] & none/32 & 3/3\\
Mistral & M & RB & 3 & 8886 & 1984 & 8326 & 1847 & 56143 & +493 [$-$421, +1408] & 1.98 [1.13, 2.83] & none/none & 2/3\\
Mistral & M & LR & 3 & 8896 & 1995 & 833 & 1847 & 48670 & +493 [$-$422, +1409] & 1.72 [0.92, 2.52] & none/none & 2/3\\
Mistral & M & TF & 3 & 8888 & 750 & 836 & 641 & 23295 & $-$713 [$-$735, $-$691] & 0.82 [0.79, 0.84] & 13/13 & 3/3\\
Mistral & M & RRTF & 3 & 0 & 515 & 0 & 436 & 9241 & $-$918 [$-$969, $-$867] & 0.32 [0.21, 0.43] & 1/0 & 3/3\\
Mistral & L & RR & 5 & 0 & 4525 & 0 & 4445 & 93433 & --- & 1.00 & --- & 3/5\\
Mistral & L & C & 5 & 13885 & 1573 & 0 & 1499 & 45443 & $-$2946 [$-$3759, $-$2133] & 0.49 [0.32, 0.65] & 4/4 & 3/5\\
Mistral & L & RB & 5 & 16920 & 2424 & 16408 & 2332 & 82400 & $-$2113 [$-$3110, $-$1116] & 0.89 [0.66, 1.11] & 15/15 & 2/5\\
Mistral & L & LR & 5 & 17303 & 2404 & 929 & 2295 & 66537 & $-$2150 [$-$3120, $-$1181] & 0.71 [0.50, 0.93] & 8/7 & 2/5\\
Mistral & L & TF & 5 & 16957 & 839 & 971 & 729 & 33342 & $-$3717 [$-$4028, $-$3405] & 0.36 [0.31, 0.41] & 4/4 & 2/5\\
Mistral & L & RRTF & 5 & 0 & 574 & 0 & 494 & 10446 & $-$3952 [$-$4203, $-$3701] & 0.11 [0.09, 0.14] & 1/0 & 4/5\\
Mistral & XL & RR & 6 & 0 & 10201 & 0 & 10130 & 212794 & --- & 1.00 & --- & 3/6\\
Mistral & XL & C & 6 & 19996 & 1528 & 0 & 1470 & 50918 & $-$8660 [$-$9244, $-$8076] & 0.24 [0.17, 0.31] & 2/1 & 4/6\\
Mistral & XL & RB & 6 & 23284 & 2416 & 22726 & 2304 & 94506 & $-$7826 [$-$8597, $-$7054] & 0.44 [0.34, 0.55] & 5/5 & 5/6\\
Mistral & XL & LR & 6 & 23265 & 2410 & 959 & 2289 & 72412 & $-$7841 [$-$8599, $-$7083] & 0.34 [0.25, 0.43] & 3/2 & 4/6\\
Mistral & XL & TF & 6 & 23268 & 827 & 957 & 723 & 39508 & $-$9407 [$-$9636, $-$9177] & 0.19 [0.16, 0.21] & 2/2 & 5/6\\
Mistral & XL & RRTF & 6 & 0 & 561 & 0 & 481 & 10175 & $-$9649 [$-$9862, $-$9436] & 0.05 [0.04, 0.06] & 1/0 & 5/6\\
\bottomrule
\end{tabular}

\end{table}
\end{landscape}
\begin{table}[!tbp]\centering
\caption{Length sweep, pre-specified cached-input pricing (pipeline abbreviations as in Table~\ref{tab:lengthpred}): cumulative tokens at $k=20$ relative to re-reading when input tokens that repeat the previous call's prefix cost $c$ of an uncached token. Ratios of mean totals (not means of paired ratios), so the no-cache column can differ from Table~\ref{tab:lengthfull} in the last digit. Mistral's cached ratios ($c=0.5$, $0.25$, $0.1$) are not shown (---): the program counted Mistral's repeated prefixes with the Qwen2.5 tokenizer, which favours memory, so those values are invalid and no claim uses them (App.~\ref{app:length}). All other cells are as written by the hashed program.}\label{tab:lengthcached}
\footnotesize\setlength{\tabcolsep}{3pt}\begin{tabular}{@{}lllrrrr@{}}\toprule
model & level & pipeline & no cache & $c=0.5$ & $c=0.25$ & $c=0.1$\\\midrule
Qwen & S & RR & 1.00 & 1.00 & 1.00 & 1.00\\
Qwen & S & C & 1.42 & 1.44 & 1.46 & 1.48\\
Qwen & S & RB & 2.40 & 2.54 & 2.68 & 2.86\\
Qwen & S & LR & 2.21 & 2.26 & 2.32 & 2.39\\
Qwen & S & TF & 1.30 & 1.46 & 1.65 & 1.88\\
Qwen & S & RRTF & 0.72 & 0.77 & 0.84 & 0.91\\
Qwen & M & RR & 1.00 & 1.00 & 1.00 & 1.00\\
Qwen & M & C & 1.63 & 1.71 & 1.83 & 2.00\\
Qwen & M & RB & 2.65 & 2.99 & 3.47 & 4.19\\
Qwen & M & LR & 2.40 & 2.59 & 2.86 & 3.27\\
Qwen & M & TF & 0.79 & 1.05 & 1.41 & 1.95\\
Qwen & M & RRTF & 0.33 & 0.39 & 0.46 & 0.58\\
Qwen & L & RR & 1.00 & 1.00 & 1.00 & 1.00\\
Qwen & L & C & 1.08 & 1.19 & 1.37 & 1.71\\
Qwen & L & RB & 1.69 & 2.04 & 2.60 & 3.65\\
Qwen & L & LR & 1.51 & 1.72 & 2.07 & 2.72\\
Qwen & L & TF & 0.37 & 0.56 & 0.88 & 1.48\\
Qwen & L & RRTF & 0.11 & 0.14 & 0.18 & 0.27\\
Qwen & XL & RR & 1.00 & 1.00 & 1.00 & 1.00\\
Qwen & XL & C & 0.59 & 0.68 & 0.83 & 1.13\\
Qwen & XL & RB & 0.89 & 1.13 & 1.53 & 2.37\\
Qwen & XL & LR & 0.78 & 0.92 & 1.17 & 1.67\\
Qwen & XL & TF & 0.20 & 0.32 & 0.53 & 0.96\\
Qwen & XL & RRTF & 0.05 & 0.06 & 0.08 & 0.13\\
Mistral & S & RR & 1.00 & 1.00 & 1.00 & 1.00\\
Mistral & S & C & 1.50 & --- & --- & ---\\
Mistral & S & RB & 2.55 & --- & --- & ---\\
Mistral & S & LR & 2.33 & --- & --- & ---\\
Mistral & S & TF & 1.38 & --- & --- & ---\\
Mistral & S & RRTF & 0.70 & --- & --- & ---\\
Mistral & M & RR & 1.00 & 1.00 & 1.00 & 1.00\\
Mistral & M & C & 1.08 & --- & --- & ---\\
Mistral & M & RB & 1.97 & --- & --- & ---\\
Mistral & M & LR & 1.71 & --- & --- & ---\\
Mistral & M & TF & 0.82 & --- & --- & ---\\
Mistral & M & RRTF & 0.32 & --- & --- & ---\\
Mistral & L & RR & 1.00 & 1.00 & 1.00 & 1.00\\
Mistral & L & C & 0.49 & --- & --- & ---\\
Mistral & L & RB & 0.88 & --- & --- & ---\\
Mistral & L & LR & 0.71 & --- & --- & ---\\
Mistral & L & TF & 0.36 & --- & --- & ---\\
Mistral & L & RRTF & 0.11 & --- & --- & ---\\
Mistral & XL & RR & 1.00 & 1.00 & 1.00 & 1.00\\
Mistral & XL & C & 0.24 & --- & --- & ---\\
Mistral & XL & RB & 0.44 & --- & --- & ---\\
Mistral & XL & LR & 0.34 & --- & --- & ---\\
Mistral & XL & TF & 0.19 & --- & --- & ---\\
Mistral & XL & RRTF & 0.05 & --- & --- & ---\\
\bottomrule
\end{tabular}

\end{table}
\begin{table}[!tbp]\centering
\caption{Length sweep, remaining pre-specified endpoints: cumulative tokens relative to re-reading at $k=1$, 5, 10 and 20 (means of per-patient paired ratios), strict completion with exact Clopper--Pearson (CP) 95\,\% intervals, and trajectories whose uses reached the commit gate (committed); RR, TF and RRTF as in Table~\ref{tab:lengthpred}.}\label{tab:lengthendpoints}
\footnotesize\setlength{\tabcolsep}{3pt}\begin{tabular}{@{}lllrrrrrlr@{}}\toprule
model & level & pipeline & $n$ & ratio@1 & ratio@5 & ratio@10 & ratio@20 & strict [95\,\% CP] & committed\\\midrule
Qwen & S & RR & 8 & --- & --- & --- & --- & 6/8 [0.35--0.97] & 8/8\\
Qwen & S & cache & 8 & 2.66 & 1.76 & 1.54 & 1.42 & 4/8 [0.16--0.84] & 8/8\\
Qwen & S & rebuild & 8 & 6.71 & 3.57 & 2.83 & 2.40 & 7/8 [0.47--1.00] & 8/8\\
Qwen & S & local repair & 8 & 4.86 & 2.93 & 2.47 & 2.21 & 7/8 [0.47--1.00] & 8/8\\
Qwen & S & TF & 8 & 3.98 & 2.03 & 1.56 & 1.30 & 8/8 [0.63--1.00] & 8/8\\
Qwen & S & RRTF & 8 & 0.73 & 0.72 & 0.72 & 0.72 & 8/8 [0.63--1.00] & 8/8\\
Qwen & M & RR & 3 & --- & --- & --- & --- & 3/3 [0.29--1.00] & 3/3\\
Qwen & M & cache & 3 & 3.32 & 2.08 & 1.80 & 1.63 & 3/3 [0.29--1.00] & 3/3\\
Qwen & M & rebuild & 3 & 7.98 & 4.08 & 3.18 & 2.66 & 3/3 [0.29--1.00] & 3/3\\
Qwen & M & local repair & 3 & 5.43 & 3.21 & 2.70 & 2.41 & 3/3 [0.29--1.00] & 3/3\\
Qwen & M & TF & 3 & 3.85 & 1.61 & 1.09 & 0.79 & 3/3 [0.29--1.00] & 3/3\\
Qwen & M & RRTF & 3 & 0.34 & 0.33 & 0.33 & 0.33 & 3/3 [0.29--1.00] & 3/3\\
Qwen & L & RR & 5 & --- & --- & --- & --- & 5/5 [0.48--1.00] & 5/5\\
Qwen & L & cache & 5 & 2.37 & 1.42 & 1.21 & 1.08 & 4/5 [0.28--0.99] & 5/5\\
Qwen & L & rebuild & 5 & 5.34 & 2.65 & 2.04 & 1.69 & 5/5 [0.48--1.00] & 5/5\\
Qwen & L & local repair & 5 & 3.53 & 2.04 & 1.70 & 1.51 & 5/5 [0.48--1.00] & 5/5\\
Qwen & L & TF & 5 & 2.38 & 0.90 & 0.56 & 0.37 & 5/5 [0.48--1.00] & 5/5\\
Qwen & L & RRTF & 5 & 0.12 & 0.11 & 0.11 & 0.11 & 5/5 [0.48--1.00] & 5/5\\
Qwen & XL & RR & 6 & --- & --- & --- & --- & 4/6 [0.22--0.96] & 6/6\\
Qwen & XL & cache & 6 & 1.50 & 0.83 & 0.68 & 0.59 & 5/6 [0.36--1.00] & 6/6\\
Qwen & XL & rebuild & 6 & 3.23 & 1.51 & 1.12 & 0.89 & 6/6 [0.54--1.00] & 6/6\\
Qwen & XL & local repair & 6 & 2.03 & 1.11 & 0.90 & 0.78 & 6/6 [0.54--1.00] & 6/6\\
Qwen & XL & TF & 6 & 1.45 & 0.53 & 0.32 & 0.20 & 6/6 [0.54--1.00] & 6/6\\
Qwen & XL & RRTF & 6 & 0.05 & 0.05 & 0.05 & 0.05 & 6/6 [0.54--1.00] & 6/6\\
\midrule
Mistral & S & RR & 8 & --- & --- & --- & --- & 4/8 [0.16--0.84] & 8/8\\
Mistral & S & cache & 8 & 2.96 & 1.90 & 1.65 & 1.51 & 5/8 [0.24--0.91] & 8/8\\
Mistral & S & rebuild & 8 & 7.35 & 3.88 & 3.05 & 2.56 & 4/8 [0.16--0.84] & 8/8\\
Mistral & S & local repair & 8 & 5.29 & 3.15 & 2.64 & 2.34 & 5/8 [0.24--0.91] & 8/8\\
Mistral & S & TF & 8 & 4.37 & 2.20 & 1.68 & 1.38 & 5/8 [0.24--0.91] & 8/8\\
Mistral & S & RRTF & 8 & 0.71 & 0.71 & 0.70 & 0.70 & 7/8 [0.47--1.00] & 8/8\\
Mistral & M & RR & 3 & --- & --- & --- & --- & 3/3 [0.29--1.00] & 3/3\\
Mistral & M & cache & 3 & 3.14 & 1.63 & 1.29 & 1.09 & 3/3 [0.29--1.00] & 3/3\\
Mistral & M & rebuild & 3 & 7.67 & 3.50 & 2.54 & 1.98 & 2/3 [0.09--0.99] & 3/3\\
Mistral & M & local repair & 3 & 4.94 & 2.58 & 2.03 & 1.72 & 2/3 [0.09--0.99] & 3/3\\
Mistral & M & TF & 3 & 4.04 & 1.68 & 1.13 & 0.82 & 3/3 [0.29--1.00] & 3/3\\
Mistral & M & RRTF & 3 & 0.34 & 0.33 & 0.32 & 0.32 & 3/3 [0.29--1.00] & 3/3\\
Mistral & L & RR & 5 & --- & --- & --- & --- & 3/5 [0.15--0.95] & 5/5\\
Mistral & L & cache & 5 & 1.90 & 0.86 & 0.62 & 0.49 & 3/5 [0.15--0.95] & 5/5\\
Mistral & L & rebuild & 5 & 4.26 & 1.78 & 1.21 & 0.89 & 2/5 [0.05--0.85] & 5/5\\
Mistral & L & local repair & 5 & 2.56 & 1.20 & 0.89 & 0.71 & 2/5 [0.05--0.85] & 5/5\\
Mistral & L & TF & 5 & 2.18 & 0.84 & 0.53 & 0.36 & 2/5 [0.05--0.85] & 5/5\\
Mistral & L & RRTF & 5 & 0.12 & 0.11 & 0.11 & 0.11 & 4/5 [0.28--0.99] & 5/5\\
Mistral & XL & RR & 6 & --- & --- & --- & --- & 3/6 [0.12--0.88] & 6/6\\
Mistral & XL & cache & 6 & 1.13 & 0.47 & 0.32 & 0.24 & 4/6 [0.22--0.96] & 6/6\\
Mistral & XL & rebuild & 6 & 2.50 & 0.99 & 0.64 & 0.44 & 5/6 [0.36--1.00] & 6/6\\
Mistral & XL & local repair & 6 & 1.42 & 0.63 & 0.44 & 0.34 & 4/6 [0.22--0.96] & 6/6\\
Mistral & XL & TF & 6 & 1.27 & 0.47 & 0.29 & 0.19 & 5/6 [0.36--1.00] & 6/6\\
Mistral & XL & RRTF & 6 & 0.05 & 0.05 & 0.05 & 0.05 & 5/6 [0.36--1.00] & 6/6\\
\bottomrule\end{tabular}

\end{table}
\begin{table}[!tbp]\centering
\caption{Length sweep, exploratory extraction diagnostic: ingest and revise extraction calls of cache, rebuild, local repair and TF combined, calls that stopped at the 2{,}048-token output cap, and outputs that failed to parse after the labelled repair, for the original and the added documents.}\label{tab:lengthextract}
\footnotesize\setlength{\tabcolsep}{4pt}\begin{tabular}{@{}llrrrrrr@{}}\toprule
 & & \multicolumn{3}{c}{original documents (D1--D4)} & \multicolumn{3}{c}{added documents}\\\cmidrule(lr){3-5}\cmidrule(l){6-8}
model & level & calls & at cap & parse errors & calls & at cap & parse errors\\\midrule
Qwen & S & 152 & 0 & 0 & 0 & 0 & 0\\
Qwen & M & 57 & 0 & 0 & 55 & 0 & 0\\
Qwen & L & 95 & 0 & 0 & 105 & 35 & 0\\
Qwen & XL & 114 & 0 & 0 & 155 & 75 & 0\\
\midrule
Mistral & S & 152 & 0 & 0 & 0 & 0 & 0\\
Mistral & M & 57 & 0 & 0 & 55 & 0 & 0\\
Mistral & L & 95 & 0 & 0 & 105 & 55 & 9\\
Mistral & XL & 114 & 0 & 0 & 155 & 60 & 35\\
\midrule
\multicolumn{2}{@{}l}{total} & 836 & 0 & 0 & 630 & 225 & 44\\
\bottomrule\end{tabular}

\end{table}

\FloatBarrier
\section{Development-stage supplement: one 32B model (EXP-L1, EXP-L2, EXP-L3)}\label{app:largemodel}\suppressfloats[t]
The supplement was run after the 7B results were known, under a pre-specification plan (hashed before any model call under it) and the same frozen build, with the same prompts, scenarios, pipelines and decoding as in Table~\ref{tab:cost}. The model is Qwen2.5-32B-Instruct (bf16) \citep{yang2024qwen25}, served by vLLM \citep{kwon2023efficient} on one 80--96 GB data-centre GPU per job. A serving check (the extraction prompt on one development document, strict JSON parse, byte-identical greedy replay) preceded the batch and is not a result. One job, 3{,}140 GPU-seconds; the eight development screening patients $\times$ 2 sizes $\times$ 4 pipelines = 64 E1 trajectories with two later uses as in Table~\ref{tab:cost}; 0 execution failures. Hardware amendment (dated before the first 32B call): the plan named an 80 GB accelerator; site routing placed the single-GPU job on a 96 GB accelerator of the same class, with model, dtype, context length and decoding unchanged. Counts are descriptive means over development patients; no test was pre-specified and the run is a third model column, not a replicate. Stale commits were 0/8 and strict completion 7/8 in all eight cells; every memory pipeline's later-use tokens exceeded re-reading's in all six memory cells (Table~\ref{tab:L1}). The revocation batch (L2) and the replacement batch (L3) of the same plan are reported below.
\begin{table}[!tbp]\centering
\caption{32B supplement (E1, eight development patients per cell): mean prompt+completion tokens per trajectory by phase (use$_0$ = initial use before the revision), model calls, derived nodes invalidated / recomputed at the revision (as in Table~\ref{tab:cost}), and total relative to re-reading at the same size. Phase means are rounded separately, so phases can differ from the total by one token.}\label{tab:L1}
\footnotesize\setlength{\tabcolsep}{3pt}\begin{tabular}{@{}llrrrrrrrr@{}}\toprule
setting & pipeline & ingest & use$_0$ & revise & later uses & total & calls & inv./rebuilt & $\times$ re-read\\\midrule
Qwen2.5-32B small & re-read & 0 & 699 & 0 & 1162 & 1861 & 3.0 & --- & 1.00\\
Qwen2.5-32B small & cache & 1878 & 853 & 0 & 1499 & 4230 & 7.0 & 1/0 & 2.27\\
Qwen2.5-32B small & rebuild & 3433 & 1286 & 2961 & 2304 & 9984 & 19.0 & 3/7 & 5.36\\
Qwen2.5-32B small & local repair & 3428 & 1279 & 516 & 2301 & 7524 & 13.0 & 3/1 & 4.04\\
\midrule
Qwen2.5-32B large & re-read & 0 & 1598 & 0 & 2976 & 4575 & 3.0 & --- & 1.00\\
Qwen2.5-32B large & cache & 6098 & 2545 & 0 & 4904 & 13546 & 10.8 & 1/0 & 2.96\\
Qwen2.5-32B large & rebuild & 9754 & 3477 & 9389 & 6899 & 29519 & 34.0 & 3/14.5 & 6.45\\
Qwen2.5-32B large & local repair & 9869 & 3592 & 1085 & 6850 & 21396 & 20.5 & 3/1 & 4.68\\
\bottomrule
\end{tabular}

\end{table}
\paragraph{Replacement with the 32B model (EXP-L3).} Development-stage supplement, not a held-out replication: run after the 7B results of batches C and C2 were known, under the same pre-specification plan and frozen build, with the frozen batch-C and C2 scenario files, the five batch-C pipelines and the frozen field-level scorer. Three cells, never pooled, each disclosing its reuse of patients: \emph{dev13} = the 13 batch-C development patients (earlier model runs on these records: batches C and E for all 13 and batches A and B for 7 of them; the third run with this replacement design); \emph{calib8} = the 8 calibration patients of C2 stratum A (second use); \emph{sealed11} = the 11 sealed patients of C2 stratum B (third use). One 80--96\,GB data-centre GPU per cell, 1{,}859, 1{,}259 and 1{,}811 GPU-seconds; 65 + 40 + 55 = 160 trajectories, 0 execution failures. Execution amendment (dated before the jobs started): the batch was first submitted as one job (default queue), cancelled before starting when the scheduler gave no start estimate, and resubmitted unchanged as one job per cell to a short-job queue. Per cell, the one-sided exact test of re-reading $>$ local repair on adoption (the C2 primary): dev13 discordant 9/0, $p=0.0020$ (rejected at $\alpha=0.05$); calib8 4/0, $p=0.0625$ (not rejected; at $n=8$ rejection needs five or more discordant pairs all on one side, or an 8-pair 7:1 split); sealed11 5/0, $p=0.031$ (rejected). A post-hoc Holm-3 correction across the three cells leaves only dev13 rejected (adjusted $p=0.006$, 0.0625, 0.0625). Smallest Holm-3 adjusted secondary $p$: 0.0015, 0.094, 0.094. No-propagation kept the superseded value in 3/13, 4/8 and 4/11. Every non-adoption again occurred with the new value visible in the use prompt; no mechanism attribution is made. Table~\ref{tab:L3} gives the per-cell counts.
\begin{table}[!tbp]\centering
\caption{32B replacement supplement (E2; frozen field-level scorer): per-cell counts for the five batch-C pipelines; dev13, calib8 and sealed11 = 13 development, 8 calibration and 11 sealed patients; no-prop.\ = no-propagation. Adopted = new strength value listed; old kept = superseded value listed; others ret.\ = every other valid medication identity retained; success = adopted $\wedge$ others retained; success$_{\text{strict}}$ additionally excludes unsupported identities. Cells are never pooled.}\label{tab:L3}
\footnotesize\begin{tabular}{@{}llrrrrr@{}}\toprule
cell & pipeline & adopted & old kept & others ret. & success & success$_{\text{strict}}$\\\midrule
dev13 & no-prop. & 0/13 & 3/13 & 12/13 & 0/13 & 0/13\\
dev13 & quarantine & 0/13 & 0/13 & 13/13 & 0/13 & 0/13\\
dev13 & re-read & 13/13 & 0/13 & 13/13 & 13/13 & 13/13\\
dev13 & rebuild & 2/13 & 0/13 & 12/13 & 2/13 & 2/13\\
dev13 & local repair & 4/13 & 0/13 & 13/13 & 4/13 & 4/13\\
\midrule
calib8 & no-prop. & 0/8 & 4/8 & 8/8 & 0/8 & 0/8\\
calib8 & quarantine & 0/8 & 0/8 & 8/8 & 0/8 & 0/8\\
calib8 & re-read & 8/8 & 0/8 & 8/8 & 8/8 & 8/8\\
calib8 & rebuild & 3/8 & 0/8 & 8/8 & 3/8 & 3/8\\
calib8 & local repair & 4/8 & 0/8 & 8/8 & 4/8 & 4/8\\
\midrule
sealed11 & no-prop. & 0/11 & 4/11 & 11/11 & 0/11 & 0/11\\
sealed11 & quarantine & 0/11 & 0/11 & 11/11 & 0/11 & 0/11\\
sealed11 & re-read & 11/11 & 0/11 & 11/11 & 11/11 & 11/11\\
sealed11 & rebuild & 6/11 & 0/11 & 11/11 & 6/11 & 6/11\\
sealed11 & local repair & 6/11 & 0/11 & 11/11 & 6/11 & 6/11\\
\bottomrule
\end{tabular}

\end{table}
\paragraph{Revocation with the 32B model (EXP-L2).} Development-stage supplement, not a held-out replication: the held-out design of \S\ref{sec:revoke} (E0, E1, E1b, E4; small size; re-read, cache, rebuild, local repair) run with Qwen2.5-32B-Instruct after the 7B results were known, under the same pre-specification plan and frozen build, graded by the frozen medication grade recorded at run time. Two strata, never pooled, each disclosing its reuse of patients: \emph{A} = 6 calibration patients never used by any model run before (scenarios generated with the frozen builder, seed 20260918; only six generic-eligible calibration patients remained after the C2 run, so the pre-specified stratum is small and its interval, not its $p$-value, is the informative summary); \emph{B} = the 20 sealed patients of the held-out run, all reused: counting batches in order of completion, this was the fourth evaluation batch for 11 of them (held-out run, C2 stratum B, the L3 sealed cell, then L2) and the second for the other 9 (held-out run, then L2); scenario files byte-identical to the held-out run's. One 80--96\,GB data-centre GPU per condition, each covering both strata, 1{,}849, 2{,}505, 2{,}479 and 2{,}020 GPU-seconds (8{,}853 in total); 26 patients $\times$ 4 conditions $\times$ 4 pipelines = 416 trajectories, 0 execution failures. Execution amendment (dated before the resubmission): the batch was first submitted as one job, cancelled before start when the scheduler gave no start estimate, and split by condition into four jobs; those four failed after about 200 seconds each (813 GPU-seconds, no model output for any scenario) because of a launcher path error; the launcher was corrected, the build, scenarios, clients and decoding were unchanged and re-verified, and the four jobs were resubmitted. Results (Table~\ref{tab:L2}): stale commits 0 in all eight E1 cells (stale is not defined in E0, E1b or E4); the still-supported drug retained in 6/6 in every stratum-A cell and 20/20 in every stratum-B cell except the cache (19/20); the non-allowed E4 drug listed by no pipeline in stratum A and by rebuild and local repair in 1/20 each in stratum B (re-reading and cache 0/20; one-sided exact McNemar $p=0.5$ for each, $p=1$ for the cache); strict completion 6/6 (re-reading) and 5/6 (memory pipelines) in stratum A and 16--18/20 in stratum B; re-reading cheapest in every cell.
\begin{table}[!tbp]\centering
\caption{32B revocation supplement (small size; frozen medication grade): per-stratum, per-condition counts for the four pipelines (stratum A = 6 calibration, B = 20 sealed patients; cond.\ = condition). committed = uses that reached the commit gate; strict = listed generic-key set equals the expected set; stale = unique-support-revoked key listed (E1); retained = still-supported key present (E1b); unsupported = the non-allowed E4 drug listed; --- = endpoint not defined in that condition; tokens = mean prompt+completion per trajectory. Strata are never pooled.}\label{tab:L2}
\footnotesize\setlength{\tabcolsep}{3pt}\begin{tabular}{@{}llllrrrrrr@{}}\toprule
stratum & cond. & pipeline & n & committed & strict & stale & retained & unsupported & tokens\\\midrule
A & E0 & re-read & 6 & 6 & 6 & --- & --- & --- & 1612\\
A & E0 & cache & 6 & 6 & 5 & --- & --- & --- & 3672\\
A & E0 & rebuild & 6 & 6 & 5 & --- & --- & --- & 5910\\
A & E0 & local repair & 6 & 6 & 5 & --- & --- & --- & 5908\\
A & E1 & re-read & 6 & 6 & 6 & 0 & --- & --- & 1675\\
A & E1 & cache & 6 & 6 & 5 & 0 & --- & --- & 3985\\
A & E1 & rebuild & 6 & 6 & 5 & 0 & --- & --- & 9207\\
A & E1 & local repair & 6 & 6 & 5 & 0 & --- & --- & 6925\\
A & E1b & re-read & 6 & 6 & 6 & --- & 6/6 & --- & 1647\\
A & E1b & cache & 6 & 6 & 5 & --- & 6/6 & --- & 3920\\
A & E1b & rebuild & 6 & 6 & 5 & --- & 6/6 & --- & 9091\\
A & E1b & local repair & 6 & 6 & 5 & --- & 6/6 & --- & 6806\\
A & E4 & re-read & 6 & 6 & 6 & --- & --- & 0 & 1748\\
A & E4 & cache & 6 & 6 & 5 & --- & --- & 0 & 4021\\
A & E4 & rebuild & 6 & 6 & 5 & --- & --- & 0 & 6545\\
A & E4 & local repair & 6 & 6 & 5 & --- & --- & 0 & 6534\\
\midrule
B & E0 & re-read & 20 & 20 & 17 & --- & --- & --- & 1762\\
B & E0 & cache & 20 & 20 & 16 & --- & --- & --- & 3842\\
B & E0 & rebuild & 20 & 20 & 16 & --- & --- & --- & 6219\\
B & E0 & local repair & 20 & 20 & 16 & --- & --- & --- & 6179\\
B & E1 & re-read & 20 & 20 & 17 & 0 & --- & --- & 1834\\
B & E1 & cache & 20 & 20 & 16 & 0 & --- & --- & 4170\\
B & E1 & rebuild & 20 & 20 & 16 & 0 & --- & --- & 9697\\
B & E1 & local repair & 20 & 20 & 16 & 0 & --- & --- & 7278\\
B & E1b & re-read & 20 & 20 & 17 & --- & 20/20 & --- & 1808\\
B & E1b & cache & 20 & 20 & 16 & --- & 19/20 & --- & 4130\\
B & E1b & rebuild & 20 & 20 & 16 & --- & 20/20 & --- & 9603\\
B & E1b & local repair & 20 & 20 & 16 & --- & 20/20 & --- & 7197\\
B & E4 & re-read & 20 & 20 & 18 & --- & --- & 0 & 1908\\
B & E4 & cache & 20 & 20 & 16 & --- & --- & 0 & 4203\\
B & E4 & rebuild & 20 & 20 & 17 & --- & --- & 1 & 6854\\
B & E4 & local repair & 20 & 20 & 16 & --- & --- & 1 & 6858\\
\bottomrule
\end{tabular}

\end{table}

\end{document}